\newif\ifJOURNAL
\JOURNALfalse
\newif\ifCONF
\CONFfalse
\newif\ifarXiv
\arXivfalse
\newif\ifWP
\WPfalse
\newif\ifFULL
\FULLfalse

\arXivtrue

\newif\ifTR		
\TRfalse
\ifarXiv\TRtrue\fi
\ifWP\TRtrue\fi
\newif\ifnotTR
\notTRtrue
\ifarXiv\notTRfalse\fi
\ifWP\notTRfalse\fi

\newif\ifnotCONF	
\notCONFtrue
\ifCONF\notCONFfalse\fi

\newif\ifnotarXiv	
\notarXivtrue
\ifarXiv\notarXivfalse\fi

\ifCONF
  \newcommand{\GTPVIII}{vovk/etal:2005AIStatslocal}
  
  \newcommand{\GTPXI}{GTP11arXiv}
  
  \newcommand{\GTPXIII}{vovk:2005ALT-GTP13}
  \newcommand{\GTPXIV}{vovk:2005ALT-GTP14}
  \newcommand{\GTPXVI}{GTP16arXiv}
\fi
\ifarXiv
  \newcommand{\GTPVIII}{GTP8arXiv}
  
  \newcommand{\GTPXIII}{GTP13arXiv}
  \newcommand{\GTPXIV}{GTP14arXiv}
  \newcommand{\GTPXVI}{GTP16arXiv}
\fi
\ifWP
  \newcommand{\GTPVIII}{GTP8}
  
  \newcommand{\GTPXIII}{GTP13local}
  \newcommand{\GTPXIV}{GTP14}
  \newcommand{\GTPXVI}{GTP16}
\fi
\ifFULL
  \newcommand{\GTPVIII}{GTP8arXiv}
  
  \newcommand{\GTPXIII}{GTP13arXiv}
  \newcommand{\GTPXIV}{GTP14arXiv}
  \newcommand{\GTPXVI}{GTP16arXiv}
\fi

\ifCONF
\documentclass{llncs}
\usepackage{amsmath,amsfonts,amssymb,latexsym,color,graphicx}
\newcommand{\Extra}[1]{}
\fi

\ifarXiv
\documentclass{article}
\usepackage{amsmath,amsfonts,amssymb,latexsym,color,graphicx}
\newcommand{\Extra}[1]{}
\fi

\ifWP
\documentclass[toc]{gtarticle}
\usepackage{amsmath,amsfonts,amssymb,latexsym,epsfig,color,graphicx}
\renewcommand{\Extra}[1]{#1}
\fi

\ifFULL
\documentclass{article}
\usepackage{amsmath,amsfonts,amssymb,latexsym,color,graphicx}
\newcommand{\Extra}[1]{\red{#1}}
\fi

\allowdisplaybreaks[4]

\newcommand{\red}[1]{\textcolor{red}{#1}}

\newcommand{\Vladimir}{Vladimir }
\newcommand{\DOT}{.}

\newcommand{\st}{\mathrel{\!|\!}}
\newcommand{\D}{\,\mathrm{d}}
\newcommand{\dd}{\mathrm{d}}

\newcommand{\sign}{\mathop{\rm sign}\nolimits}
\newcommand{\minsq}{\mathop{\mathrm{min}^2}\nolimits}
\newcommand{\risk}{\mathop{\mathrm{risk}}\nolimits}
\newcommand{\FS}{\mathrm{FS}}		

\newcommand{\K}{\mathcal{K}}		
\newcommand{\kkk}{\mathbf{k}}		
\newcommand{\ccc}{\mathbf{c}}		

\newcommand{\CCC}{\mathcal{C}}		
\newcommand{\FFF}{\mathcal{F}}		
\newcommand{\HHH}{\mathcal{H}}		

\newcommand{\bbbe}{\mathbb{E}}		
\newcommand{\Expect}{\mathop{\bbbe}\nolimits}

\ifnotCONF
\newcommand{\bbbr}{\mathbb{R}}		
\newtheorem{lemma}{Lemma}

\newtheorem{corollary}{Corollary}
\newtheorem{theorem}{Theorem}
\newenvironment{proof}
  {\trivlist\item[\hskip\labelsep\textbf{Proof}]}
  {\endtrivlist}
\fi

\newcommand{\boxforqed}{\rule{.3em}{1.5ex}}
\newcommand{\qedtext}{\unskip\nobreak\hfil
  \penalty50\hskip1em\null\nobreak\hfil\boxforqed
  \parfillskip=0pt\finalhyphendemerits=0\endgraf}
\newcommand{\qedmath}{\tag*{\boxforqed}}
\newenvironment{remark*}
  {\trivlist\item[\hskip\labelsep{\bfseries Remark}]\relax}
  {\endtrivlist}

\newlength{\IndentI}
\newlength{\IndentII}
\newlength{\IndentIII}
\newlength{\WidthI}
\newlength{\WidthII}
\newlength{\WidthIII}
\ifCONF
\title{On-line regression competitive with reproducing kernel Hilbert spaces (extended abstract)}
\author{Vladimir Vovk}
\institute{Computer Learning Research Centre,
  Department of Computer Science\\
  Royal Holloway, University of London,
  Egham, Surrey TW20 0EX, UK\\
  \email{vovk@cs.rhul.ac.uk}}
\fi

\ifarXiv
\title{On-line regression competitive with reproducing kernel Hilbert spaces}
\author{Vladimir Vovk\\
\texttt{vovk{\rm@}cs.rhul.ac.uk}\\
\texttt{http://vovk.net}}
\fi

\ifWP
\title{On-line regression competitive with reproducing kernel Hilbert spaces}
\author{Vladimir Vovk}

\fi

\ifFULL
\title{On-line regression competitive with reproducing kernel Hilbert spaces}
\author{Vladimir Vovk\\
\texttt{vovk{\rm@}cs.rhul.ac.uk}\\
\texttt{http://vovk.net}}
\fi

\begin{document}
\maketitle
\begin{abstract}
  We consider the problem of on-line prediction of real-valued labels,
  assumed bounded in absolute value by a known constant,
  of new objects from known labeled objects.
  The prediction algorithm's performance is measured by the squared deviation
  of the predictions from the actual labels.
  No stochastic assumptions are made about the way the labels and objects
  are generated.
  Instead, we are given a benchmark class of prediction rules
  some of which are hoped to produce good predictions.
  We show that for a wide range of infinite-dimensional benchmark classes
  one can construct a prediction algorithm
  whose cumulative loss over the first $N$ examples
  does not exceed the cumulative loss of any prediction rule in the class
  plus $O(\surd{N})$;
  the main differences from the known results
  are that we do not impose any upper bound
  on the norm of the considered prediction rules
  and that we achieve an optimal leading term
  in the excess loss of our algorithm.
  If the benchmark class is ``universal''
  (dense in the class of continuous functions on each compact set),
  this provides an on-line non-stochastic analogue of universally consistent prediction
  in non-parametric statistics.
  We use two proof techniques:
  one is based on the Aggregating Algorithm
  and the other on the recently developed method of defensive forecasting.
  \ifarXiv
    \begin{remark*}
      The main difference of the current (second) version of this technical report
      from the previous version
      is a fuller discussion of the related literature.
      The latter is, however, massive,
      and it is likely that even in this version some important related results
      are missing.
    \end{remark*}
  \fi
\end{abstract}

\section{Introduction}
\label{sec:introduction}

The traditional, and still dominant, approach to the problem of regression
is statistical:
the objects and their real-valued labels
are assumed to be generated independently
from the same probability distribution,
and a typical goal is to find a prediction rule
with a small expected loss.
A newer approach is ``competitive on-line regression'',
in which
the goal is to perform almost as well as the best rules
in a given benchmark class of prediction rules.
(See, e.g., \cite{kivinen/warmuth:1997}, \S1, or \cite{vovk:2001competitive}, \S4,
for reviews of some relevant literature.)
Unlike the statistical theory of regression,
no stochastic assumptions are made about the data.

A great impetus for the development of the statistical theories of regression
and pattern recognition
(see, e.g., \cite{gyorfi/etal:2002} and, especially,
\cite{devroye/etal:1996}, Preface and Chapter 1)
has been Stone's 1977 result \cite{stone:1977} that there exists
a ``universally consistent'' prediction algorithm:
an algorithm that asymptotically achieves,
with probability one (or high probability),
the best possible expected loss.
The property of universal consistency is very attractive,
but it is asymptotic and does not tell us anything about finite data sequences.
Stone's result provided a direction
in which more practicable results have been sought.

Surprisingly,
it appears that universal consistency
has not been even defined in competitive on-line learning theory.
We propose such a definition in \S\ref{sec:main};
in \S\ref{sec:review} we will see how close papers such as
\cite{cesabianchi/long/warmuth:1996,auer/etal:2002}
came to constructing universally consistent algorithms.
However, our Corollary \ref{cor:universal} in \S\ref{sec:main}
appears to be the first explicit statement
about the existence of the latter.

As in the case of statistical regression,
universal consistency is only a minimal requirement;
one also wants good rates of convergence,
ideally not involving unknown constants,
for universal benchmark classes.
The notion of universality is discussed, formally and informally,
at the end of \S\ref{sec:main} and in \S\ref{sec:examples};
we will argue that universality for benchmark classes is a matter of degree.
Our main results, Theorems \ref{thm:main1}--\ref{thm:main3},
are stated in \S\ref{sec:main} and proved in \S\S\ref{sec:proof1}--\ref{sec:proof3}.
They describe properties of universality of our prediction algorithms,
some of which are described explicitly in the last section, \S\ref{sec:algorithms}.
In \S\ref{sec:PAC}, Theorem \ref{thm:main1}
is applied to the case where the objects and their labels
are drawn independently from the same distribution.
In \S\ref{sec:examples}
we consider
\ifCONF an important benchmark class \fi
\ifnotCONF some interesting benchmark classes \fi
of prediction rules,
and in \S\ref{sec:review} we compare our results to some related ones
in the literature.

In this paper
we use two very different proof techniques:
the old one introduced in \cite{vovk:1990,vovk:2001competitive}
and the one developed in \cite{\GTPXIV};
we are especially interested in the latter
since it appears much more versatile,
and competitive on-line regression is a good testing ground
to develop it.
This technique has its origin in Foster and Vohra's paper \cite{foster/vohra:1998},
which demonstrated the existence of a randomized forecasting strategy
that produces asymptotically well-calibrated forecasts
with probability one.
Foster and Vohra's result was translated into the game-theoretic foundations of probability
(see, e.g., \cite{shafer/vovk:2001}) in \cite{vovk/shafer:2005RSS}.
In June 2004 Akimichi Takemura further developed the method of \cite{vovk/shafer:2005RSS}
showing that for any continuous game-theoretic law of probability
there exists a forecasting strategy that perfectly satisfies
this law of probability;
such a strategy was called a ``defensive forecasting strategy'' in \cite{\GTPVIII}.
An important special case of defensive forecasting
is where the law of probability asserts good calibration and resolution of the forecasts;
it was explored in \cite{\GTPXIII},
where, in particular, a non-asymptotic version of Foster and Vohra's result was proved.
In \cite{\GTPXIV} it was shown that the corresponding forecasting strategies
lead to a small cumulative loss
in a fairly wide class of decision protocols.
That paper only dealt with the case of binary classification,
and in this paper similar results are proved for on-line regression.
As the loss function we use square-loss,
which leads to significant simplifications as compared with \cite{\GTPXIV}.
(Despite \cite{foster/vohra:1998} being the source of our approach,
our proof technique appears to have lost all connections
with that paper and papers,
such as \cite{lehrer:2001,sandroni:2003,sandroni/etal:2003,kakade/foster:2004},
further developing it.)

Our results are closely related to those of
Cesa-Bianchi \emph{et al.}\ \cite{cesabianchi/long/warmuth:1996}
and Auer \emph{et al.}\ \cite{auer/etal:2002},
but we postpone a detailed discussion to \S\ref{sec:review}.

\section{Main results}
\label{sec:main}

The simple perfect-information protocol of this section is:

\bigskip

\parshape=5
\IndentI  \WidthI
\IndentII \WidthII
\IndentII \WidthII
\IndentII \WidthII
\IndentI  \WidthI
\noindent
FOR $n=1,2,\dots$:\\
  Reality announces $x_n\in\mathbf{X}$.\\
  Predictor announces $\mu_n\in\bbbr$.\\
  Reality announces $y_n\in[-Y,Y]$.\\
END FOR.

\bigskip

\noindent
At the beginning of each round $n$ Predictor is shown an object $x_n$
whose label $y_n$ is to be predicted.
The set of \emph{a priori} possible objects is called the \emph{object space}
and denoted $\mathbf{X}$;
of course, we always assume $\mathbf{X}\ne\emptyset$.
After Predictor announces his prediction $\mu_n$ for the object's label
he is shown the actual label $y_n\in\bbbr$.
We assume known an \emph{a priori} upper bound $Y\in(0,\infty)$
on the absolute values of the labels $y_n$.
We will sometimes refer to pairs $(x_n,y_n)$ as \emph{examples}.
By an \emph{on-line prediction algorithm} we mean a strategy for Predictor
in this protocol;
in this paper, however, we are not concerned with computational complexity
of our prediction algorithms.

Predictor's loss on round $n$ is measured by
$(y_n-\mu_n)^2$,
and so his cumulative loss after $N$ rounds of the game is
$
  \sum_{n=1}^N
  (y_n-\mu_n)^2
$.
His goal is ``universal prediction'',
in the following, rather vague, sense.
If $D:\mathbf{X}\to\bbbr$ is a ``prediction rule''
(i.e., the function $D$ is interpreted as a rule for choosing the prediction
based on the current object),
he would like to have
\begin{equation}\label{eq:goal}
  \sum_{n=1}^N
  (y_n-\mu_n)^2
  \lessapprox
  \sum_{n=1}^N
  (y_n-D(x_n))^2
\end{equation}
($\lessapprox$ meaning ``not much greater than'')
provided $D$ is not ``too complex''.
Technically,
we will be interested in the case where the prediction rule $D$ is assumed to belong
to a large reproducing kernel Hilbert space
(to be defined shortly)
and the complexity of $D$ is measured by its norm.

As already mentioned,
the results of this section are closely related to several results
in \cite{cesabianchi/long/warmuth:1996} and \cite{auer/etal:2002};
see \S\ref{sec:review}.

\subsection*{Reproducing kernel Hilbert spaces}

A \emph{reproducing kernel Hilbert space} (RKHS) on a set $Z$
(such as $Z=\mathbf{X}$)
is a Hilbert space $\FFF$ of real-valued functions on $Z$
such that the evaluation functional $f\in\FFF\mapsto f(z)$
is continuous for each $z\in Z$.
We will use the notation $\ccc_{\FFF}(z)$ for the norm of this functional:
\begin{equation*}
  \ccc_{\FFF}(z)
  :=
  \sup_{f:\left\|f\right\|_{\FFF}\le1}
  \left|
    f(z)
  \right|.
\end{equation*}
Let 
\begin{equation}\label{eq:C}
  \ccc_{\FFF}
  :=
  \sup_{z\in Z}
  \ccc_{\FFF}(z);
\end{equation}
we will be interested in the case $\ccc_{\FFF}<\infty$.

Examples of RKHS will be given in \S\ref{sec:examples}.

\subsection*{Main theorems}

Suppose Predictor's goal is to compete with prediction rules $D$
from an RKHS $\FFF$ on $\mathbf{X}$.
The three theorems that we state in this subsection
bound the difference between the left-hand and right-hand sides of (\ref{eq:goal});
this bound will be called the \emph{regret term}.
The simplest regret term, given in the first theorem,
is in terms of $\ccc_{\FFF}$, $\left\|D\right\|_{\FFF}$, and $N$.
\begin{theorem}\label{thm:main1}
  Let $\FFF$ be an RKHS on $\mathbf{X}$.
  There exists an on-line prediction algorithm producing $\mu_n\in[-Y,Y]$
  that are guaranteed to satisfy
  \begin{equation}\label{eq:goal1}
    \sum_{n=1}^N
    (y_n-\mu_n)^2
    \le
    \sum_{n=1}^N
    (y_n-D(x_n))^2
    +
    2
    Y
    \sqrt{\ccc_{\FFF}^2 + 1}
    \left(
      \left\|D\right\|_{\FFF} + Y
    \right)
    \sqrt{N}
  \end{equation}
  for all $N=1,2,\ldots$ and all $D\in\FFF$.
\end{theorem}
The regret term in the second theorem is in terms of
$\ccc_{\FFF}$, $\left\|D\right\|_{\FFF}$, and the cumulative loss of $D$
(which can be significantly less than $N$).
\begin{theorem}\label{thm:main2}
  Let $\FFF$ be an RKHS on $\mathbf{X}$.
  There exists an on-line prediction algorithm producing $\mu_n\in[-Y,Y]$
  that are guaranteed to satisfy
  \begin{multline}\label{eq:goal2}
    \sum_{n=1}^N
    (y_n-\mu_n)^2
    \le
    \sum_{n=1}^N
    (y_n-D(x_n))^2\\
    +
    2
    \sqrt{\ccc_{\FFF}^2 + 1}
    \left(
      \left\|D\right\|_{\FFF}
      +
      Y
    \right)
    \sqrt
    {
      \sum_{n=1}^N
      (y_n-D(x_n))^2
      +
      \left(
        \ccc_{\FFF}^2
        +
        1
      \right)
      \left(
        \left\|D\right\|_{\FFF}
        +
        Y
      \right)^2
    }\\
    +
    2
    \left(
      \ccc_{\FFF}^2 + 1
    \right)
    \left(
      \left\|D\right\|_{\FFF}
      +
      Y
    \right)^2
  \end{multline}
  for all $N$ and all $D\in\FFF$.
\end{theorem}
The regret term of Theorem~\ref{thm:main2} is close to being stronger
than that of Theorem~\ref{thm:main1}:
the former is at most twice as large as the latter plus an additive constant,
if we restrict our attention to the prediction rules $D$
such that $\left\|D\right\|_{\FFF}$ is bounded by a constant
and $\left|D(x)\right|\le Y$, $\forall x\in\mathbf{X}$.

On-line prediction algorithms achieving (\ref{eq:goal1}) and (\ref{eq:goal2})
will be stated explicitly in \S\ref{sec:algorithms}.
They are based on the idea of defensive forecasting.
However, the regression problem considered in this paper
is very well studied,
and one can hardly hope to beat the known techniques.
The next theorem gives an upper bound
of the regret term achievable
by using the procedure (``Aggregating Algorithm'', or AA)
described in \cite{vovk:1990}
and applied to the problem of regression in \cite{vovk:2001competitive}
and \cite{gammerman/etal:2004UAI}.
A popular alternative technique based on the gradient descent method
could also be used,
but it tends to lead to worse leading constants:
see \S\ref{sec:review} for details.
\begin{theorem}\label{thm:main3}
  Let $\FFF$ be a separable RKHS on $\mathbf{X}$.
  There exists an on-line prediction algorithm producing $\mu_n\in[-Y,Y]$
  that are guaranteed to satisfy
  \begin{multline}\label{eq:goal3}
    \sum_{n=1}^N
    (y_n-\mu_n)^2
    \le
    \sum_{n=1}^N
    (y_n-D(x_n))^2\\
    -
    2
    Y^2
    \ln
    \left(
      \Gamma
      \left(
        \frac{N}{2}+1
      \right)
      U
      \left(
        \frac{N}{2}+1,
        0,
        \frac{\ccc_{\FFF}^2\left\|D\right\|_{\FFF}^2}{2Y^2}
      \right)
    \right)\\
    \le
    \sum_{n=1}^N
    (y_n-D(x_n))^2
    +
    2
    Y
    \max
    \left(
      \ccc_{\FFF}
      \left\|D\right\|_{\FFF},
      Y
      \delta
      N^{-1/2+\delta}
    \right)
    \sqrt{N+2}\\
    +
    \frac{3}{2}
    Y^2
    \ln N
    +
    \frac{\ccc_{\FFF}^2\left\|D\right\|_{\FFF}^2}{4}
    +
    O(Y^2)
  \end{multline}
  for all $N=1,2,\ldots$ and all $D\in\FFF$,
  where $\delta>0$ is an arbitrarily small constant,
  $\Gamma$ is the gamma function (\cite{abramowitz/stegun:1964}, Chapter 6),
  and $U$ is Kummer's $U$ function (\cite{abramowitz/stegun:1964}, Chapter 13).
  The constant implicit in $O(Y^2)$
  depends only on $\delta$.
\end{theorem}

\ifFULL\bluebegin
  An alternative representation of Kummer's $U$ function
  is Whittaker's function
  \begin{equation*}
    W_{\kappa,\mu}(z)
    =
    e^{-z/2}
    z^{1/2+\mu}
    U
    \left(
      \frac12 + \mu - \kappa,
      1 + 2\mu,
      z
    \right)
  \end{equation*}
  (\cite{abramowitz/stegun:1964}, p.~505, 13.1.33).
  Therefore, the first inequality in (\ref{eq:goal3})
  can be rewritten as
  \begin{multline*}
    \sum_{n=1}^N
    (y_n-\mu_n)^2
    \le
    \sum_{n=1}^N
    (y_n-D(x_n))^2\\
    -
    2
    Y^2
    \ln
    \left(
      \Gamma
      \left(
        \frac{N}{2}+1
      \right)
      \exp
      \left(
        \frac{\ccc_{\FFF}^2\left\|D\right\|_{\FFF}^2}{4Y^2}
      \right)
      W_{-N/2-1,-1/2}
      \left(
        \frac{\ccc_{\FFF}^2\left\|D\right\|_{\FFF}^2}{2Y^2}
      \right)
    \right).
  \end{multline*}
\blueend\fi

\begin{figure}[bt]
  \centering
  \makebox{\includegraphics[width=10cm]{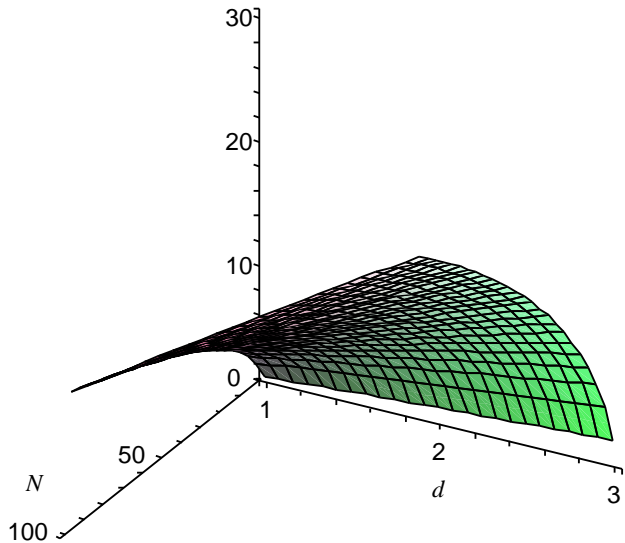}}
  \caption{\label{fig:kummer}The graph of the function
    $f(N,d)$ for $N=1,\ldots,100$ and $d\in[1,3]$.
    The two final values at the corners are $f(100,1)\approx12.37$ and $f(100,3)\approx30.15$.}
\end{figure}

The bound of Theorem~\ref{thm:main3} is even closer to being stronger
than that of Theorem~\ref{thm:main1} as $N\to\infty$:
the leading constant is the same,
$2Y\ccc_{\FFF}\left\|D\right\|_{\FFF}$
(assuming $\left\|D\right\|_{\FFF}\gg Y$ and $\ccc_{\FFF}\gg1$),
but the other terms are considerably better.
The main disadvantage of the bound (\ref{eq:goal3})
is the asymptotic character of (namely, the presence of the $O$ term in)
its more explicit version.
The version involving the gamma and Kummer's $U$ functions
is not intuitive,
but it can be evaluated using standard libraries;
the function
\begin{equation*}
  f(N,d)
  :=
  -\ln
  \left(
    \Gamma
    \left(
      \frac{N}{2}+1
    \right)
    U
    \left(
      \frac{N}{2}+1,0,\frac{d^2}{2}
    \right)
  \right)
\end{equation*}
is plotted in Figure \ref{fig:kummer}.

The condition of separability in Theorem \ref{thm:main3} does not appear restrictive;
in particular, it is satisfied for all examples considered in \S\ref{sec:examples}.

Finally, we give a lower bound
(a version of Theorem VII.2 in \cite{cesabianchi/long/warmuth:1996})
showing that the leading constant
$2Y\ccc_{\FFF}\left\|D\right\|_{\FFF}$
is optimal.
\begin{theorem}\label{thm:main4}
  Suppose the object space is $\mathbf{X}=\bbbr$.
  For any positive constant $c$
  there exists an RKHS
  \ifFULL\bluebegin
    (which can be chosen universal, in the sense of the following subsection)
  \blueend\fi
  $\FFF$ on $\mathbf{X}$
  with $\ccc_{\FFF}=c$ and a strategy for Reality
  satisfying the following property.
  For any $N=1,2,\ldots$, any positive constant $d\le(Y/\ccc_{\FFF})\surd N$,
  and any on-line prediction algorithm,
  there exists a prediction rule $D\in\FFF$
  such that $\left\|D\right\|_{\FFF}=d$ and
  \begin{equation}\label{eq:goal4}
    \sum_{n=1}^N
    (y_n-\mu_n)^2
    \ge
    \sum_{n=1}^N
    (y_n-D(x_n))^2
    +
    2Y
    \ccc_{\FFF}
    \left\|D\right\|_{\FFF}
    \sqrt{N}
    -
    \ccc_{\FFF}^2\left\|D\right\|_{\FFF}^2,
  \end{equation}
  where, as usual, $\mu_n$ are the predictions produced
  by the on-line prediction algorithm
  and $(x_n,y_n)$ are Reality's moves.
\end{theorem}

Theorems \ref{thm:main3} and \ref{thm:main4}
are proved in \S\ref{sec:proof3} and \S\ref{sec:proof4},
respectively.
From the proof of Theorem \ref{thm:main4}
it will be clear that similar lower bounds also hold
when $\mathbf{X}=\bbbr$ is replaced
by any regular (e.g., open) subset of a Euclidean space.

\begin{remark*}\label{p:remark1}
  If $\ccc_{\FFF}=\infty$
  but it is known in advance that all objects $x_n$, $n=1,2,\ldots$,
  will be chosen from a set $A\subseteq\mathbf{X}$ satisfying
  $X:=\sup_{x\in A}\ccc_{\FFF}(x)<\infty$,
  Theorem \ref{thm:main1}--\ref{thm:main4} will continue to hold
  when $\ccc_{\FFF}$ is replaced by $X$.
\end{remark*}

\subsection*{Universal consistency}

We say that an RKHS $\FFF$ on $Z$ is \emph{universal}
if $Z$ is a topological space
and for every compact subset $A$ of $Z$
every continuous function on $A$
can be arbitrarily well approximated in the metric $C(A)$
by functions in $\FFF$;
in the case of compact $Z$ this coincides
with the definition given in \cite{steinwart:2001} (Definition 4).
All examples of RKHS given in \S\ref{sec:examples} are universal.

Suppose the object space $\mathbf{X}$ is a topological space;
as in the rest of the paper, we are assuming that $\lvert y_n\rvert$
are bounded by a known constant $Y$.
Let us say that an on-line prediction algorithm is \emph{universally consistent}
if its predictions $\mu_n$ always satisfy
\begin{multline}\label{eq:consistency}
  \left(
    x_n\in A,
    \forall n\in\{1,2,\ldots\}
  \right)\\
  \Longrightarrow
  \limsup_{N\to\infty}
  \left(
    \frac1N
    \sum_{n=1}^N
    (y_n-\mu_n)^2
    -
    \frac1N
    \sum_{n=1}^N
    (y_n-D(x_n))^2
  \right)
  \le
  0
\end{multline}
for any compact subset $A$ of $\mathbf{X}$
and any continuous decision rule $D$
(cf.\ (\ref{eq:goal})).
By the Tietze--Uryson theorem
(\cite{dudley:2002}, Theorem 2.6.4 on p.~65),
if $\mathbf{X}$ is a normal topological space,
we will obtain an equivalent definition
allowing $D$ to be any continuous function from $A$ to $\bbbr$.

The definitions of this subsection are most intuitive
in the case of compact $\mathbf{X}$,
and in our informal discussion we will be making this assumption.
The main remaining difference of our definition of universal consistency
from the statistical one \cite{stone:1977}
is that we require $D$ to be continuous.
If $D$ is allowed to be discontinuous,
(\ref{eq:consistency}) is impossible to achieve:
no matter how Predictor chooses his predictions $\mu_n$,
Reality can choose
\begin{equation*}
  x_n
  :=
  \sum_{i=1}^{n-1}
  \frac{\sign(\mu_i)}{3^i},
  \quad
  y_n
  :=
  \begin{cases}
    1 & \text{if $\mu_n<0$}\\
    -1 & \text{otherwise}
  \end{cases}
\end{equation*}
(assuming $\mathbf{X}\supseteq[-1,1]$ and $Y\ge1$),
foiling (\ref{eq:consistency}) for the prediction rule
\begin{equation*}
  D(x)
  :=
  \begin{cases}
    -1 & \text{if $x<\sum_{i=1}^{\infty} \sign(\mu_i)/3^i$}\\
    1 & \text{otherwise}.
  \end{cases}
\end{equation*}
A positive argument in favor of the requirement of continuity of $D$
is that it is natural for Predictor to compete only
with computable prediction strategy,
and continuity is often regarded as a necessary condition for computability
(Brouwer's ``continuity principle'').

The existence of universal RKHS on Euclidean spaces $\bbbr^m$
(see \S\ref{sec:examples})
implies the following proposition.
\begin{corollary}\label{cor:universal}
  If $\mathbf{X}\subseteq\bbbr^m$ for some $m=1,2,\ldots$,
  there exists a universally consistent on-line prediction algorithm.
\end{corollary}
\begin{proof}
  Any on-line prediction algorithm satisfying (\ref{eq:goal1}) of Theorem \ref{thm:main1}
  for a universal RKHS $\FFF$ on $\bbbr^m$ will be universal.
  Indeed, let $A\subseteq\mathbf{X}$ be compact,
  $f$ be a continuous function on $\mathbf{X}$,
  and $\epsilon>0$.
  Suppose $x_n\in A$, $n=1,2,\ldots$\,.
  Our goal is to prove that
  \begin{equation*}
    \frac1N
    \sum_{n=1}^N
    (y_n-\mu_n)^2
    \le
    \frac1N
    \sum_{n=1}^N
    (y_n-f(x_n))^2
    +
    \epsilon
  \end{equation*}
  from some $N$ on.
  It suffices to choose $D\in\FFF$ at a distance at most $\epsilon/(8Y)$
  from $f$ in the metric $C(A)$,
  apply (\ref{eq:goal1}) to $D$,
  and notice that
  \begin{equation*}
    \left|
      \frac1N
      \sum_{n=1}^N
      (y_n-D(x_n))^2
      -
      \frac1N
      \sum_{n=1}^N
      (y_n-f(x_n))^2
    \right|
    \le
    4Y\frac{\epsilon}{8Y}
    =
    \frac{\epsilon}{2}
  \end{equation*}
  (this calculation assumes that $f$ and $D$ take values in $[-Y,Y]$;
  we can always achieve this by truncating $f$ and $D$:
  truncation does not lead outside the universal RKHS
  described in \S\ref{sec:examples}).
  \qedtext
\end{proof}

\begin{remark*}\label{p:remark2}
It is easy to extend Corollary \ref{cor:universal}
to the case where $\mathbf{X}$ is a separable metric space
or a compact metric space:
indeed, by Theorem 4.2.10 in \cite{engelking:1989}
the Hilbert cube is a universal space
for all separable metric spaces
and for all compact metric spaces,
and every continuous function on the Hilbert cube
(we are interested in continuous extensions
of continuous functions on compact subsets),
being uniformly continuous
(see, e.g., \cite{dudley:2002}, Corollary 2.4.6 on p.~52),
can be arbitrarily well approximated
by functions that only depend on the first $m$ coordinates of their argument;
it remains to notice that the on-line prediction algorithms
satisfying the condition of Theorem \ref{thm:main1}
for universal RKHS on $[0,1]^m$ can be merged into one on-line prediction algorithm
using, e.g., the Aggregating Algorithm.
\end{remark*}

So far in this subsection we have only discussed
the asymptotic notion of universal consistency,
although it is clear that one needs universality in a stronger sense.
In practical problems,
it is not enough for the benchmark class $\FFF$ to be universal;
we also want as many prediction rules $D$ as possible to belong to $\FFF$,
or at least to be well approximated by the elements of $\FFF$;
we also want $\left\|D\right\|_{\FFF}$ to be as small as possible.
The Sobolev spaces on $[0,1]^m$ discussed in \S\ref{sec:examples}
are not only universal RKHS
but also include all functions that are smooth in a fairly weak sense.
However, the Hilbert-space methods have their limitations:
it is not clear, e.g., how to apply them to functions
that are as ``smooth'' as typical trajectories of the Brownian motion.
These larger benchmark classes seem to require Banach-space methods:
see \cite{\GTPXVI}.

\section{Implications for the statistical theory of regression}
\label{sec:PAC}

So far we have not made any stochastic assumptions
about the way the examples are produced.
In this section we derive simple implications from Theorem \ref{thm:main1}
for the statistical learning framework,
assuming that the examples $(x_n,y_n)$ are drawn independently
from some probability distribution on $\mathbf{X}\times[-Y,Y]$.
Similar implications can be derived
from the results of \cite{cesabianchi/long/warmuth:1996},
\cite{auer/etal:2002},
and some other papers
(see the next section);
the corollary stated in this section, however,
has somewhat better constants.

\subsection*{Generalization bounds}

The \emph{risk} of a prediction rule $D:\mathbf{X}\to\bbbr$
with respect to a probability distribution $P$ on $\mathbf{X}\times[-Y,Y]$
is defined as
\begin{equation*}
  \risk_P(D)
  :=
  \int_{\mathbf{X}\times[-Y,Y]}
    (y-D(x))^2
  P(\dd x,\dd y).
\end{equation*}
Our goal in this section is to construct,
from a given sample,
a prediction rule whose risk
is competitive with the risk of small-norm prediction rules in a given RKHS.
As shown in \cite{cesabianchi/etal:2004}
(with similar results obtained earlier in \cite{blum/etal:1999}
and before that in \cite{littlestone:1989}),
this can be easily done once we have a competitive on-line algorithm
(such as those in Theorems \ref{thm:main1}--\ref{thm:main3}).

Fix an on-line prediction algorithm
and a sequence of examples
\begin{equation*}
  (x_1,y_1),(x_2,y_2),\ldots\,.
\end{equation*}
For each $n=1,2,\ldots$,
let $H_n:\mathbf{X}\to\bbbr$ be the function
that maps each $x\in\mathbf{X}$ to the prediction $\mu_n\in\bbbr$
output by the algorithm
when fed with $(x_1,y_1),\ldots,(x_{n-1},y_{n-1}),x$.
We will say that the prediction rule
\begin{equation*}
  \overline{H}_N(x)
  :=
  \frac1N
  \sum_{n=1}^N
  H_n(x)
\end{equation*}
is \emph{obtained by averaging} from the on-line prediction algorithm.
\begin{corollary}\label{cor:main1}
  Let $\FFF$ be an RKHS on $\mathbf{X}$,
  let $D\in\FFF$ be such that $D(x)\in[-Y,Y]$ for all $x\in\mathbf{X}$,
  and let $\overline{H}_N$, $N=1,2,\ldots$,
  be the prediction rules obtained by averaging
  from some on-line prediction algorithm guaranteeing (\ref{eq:goal1}).
  For any probability distribution $P$ on $\mathbf{X}\times[-Y,Y]$,
  any $N=1,2,\ldots$,
  and any $\delta>0$,
  \begin{equation}\label{eq:cor1}
    \risk_P(\overline{H}_N)
    \le
    \risk_P(D)
    +
    \frac{2Y}{\sqrt{N}}
    \left(
      \sqrt{\ccc_{\FFF}^2 + 1}
      \left(
        \left\|D\right\|_{\FFF} + Y
      \right)
      +
      2Y
      \sqrt{2\ln\frac{2}{\delta}}
    \right)
  \end{equation}
  with probability at least $1-\delta$.
\end{corollary}
\begin{proof}
  For a suitable choice of $\epsilon>0$,
  we will have
  \begin{align}
    \risk_P(\overline{H}_N)
    &\le
    \frac1N
    \sum_{n=1}^N
    \risk_P(H_n)
    \label{eq:jensen}\\
    &\le
    \frac1N
    \sum_{n=1}^N
    (y_n-H_n(x_n))^2
    +
    \epsilon
    \label{eq:hoeffding1}\\
    &\le
    \frac1N
    \sum_{n=1}^N
    (y_n-D(x_n))^2
    +
    \frac{2Y}{\sqrt{N}}
    \sqrt{\ccc_{\FFF}^2 + 1}
    \left(
      \left\|D\right\|_{\FFF} + Y
    \right)
    +
    \epsilon
    \label{eq:I}\\
    &\le
    \frac1N
    \sum_{n=1}^N
    \risk_P(D)
    +
    \frac{2Y}{\sqrt{N}}
    \sqrt{\ccc_{\FFF}^2 + 1}
    \left(
      \left\|D\right\|_{\FFF} + Y
    \right)
    +
    2\epsilon
    \label{eq:hoeffding2}\\
    &=
    \risk_P(D)
    +
    \frac{2Y}{\sqrt{N}}
    \sqrt{\ccc_{\FFF}^2 + 1}
    \left(
      \left\|D\right\|_{\FFF} + Y
    \right)
    +
    2\epsilon
    \notag
  \end{align}
  with probability at least $1-\delta$.
  The inequalities (\ref{eq:jensen}) and (\ref{eq:I}) always hold:
  the first follows from the convexity of the function $t\mapsto t^2$,
  and the second from Theorem~\ref{thm:main1}.
  By Hoeffding's martingale inequality
  (\cite{hoeffding:1963}, Theorem 1 and the remark at the end of \S2;
  see also \cite{devroye/etal:1996}, Theorem 9.1 on p.~135),
  (\ref{eq:hoeffding1}) and (\ref{eq:hoeffding2})
  will hold with probability at least
  $
    1
    -
    e^{-\epsilon^2N/(8Y^4)}
  $;
  to make the probability of their conjunction at least $1-\delta$,
  it suffices to find $\epsilon$ from the equation
  $
    e^{-\epsilon^2N/(8Y^4)}
    =
    \delta/2
  $,
  which gives
  \begin{equation*}
    \epsilon
    =
    \frac{2Y^2}{\sqrt{N}}
    \sqrt{2\ln\frac{2}{\delta}}.
    \qedmath
  \end{equation*}
\end{proof}

In Corollary \ref{cor:main1} \ifFULL\bluebegin and Corollary \ref{cor:main2}\blueend\fi
we only consider prediction rules taking values in $[-Y,Y]$;
this is not a real restriction if the RKHS $\FFF$
satisfies $D\in\FFF\Longrightarrow\left|D\right|\in\FFF$,
as the examples of RKHS considered in \S\ref{sec:examples} do.

\ifFULL\bluebegin
Corollary \ref{cor:main1} can be restated
using Valiant's notion of sample complexity
(see, e.g., \cite{devroye/etal:1996}, \S12.6, p.~201).
Let us say that,
for given $\epsilon>0$ and $\delta>0$,
\emph{the sample complexity of a class $\CCC$ of prediction rules
does not exceed $N$}
if it is possible to produce a prediction rule $H$ satisfying
\begin{equation}\label{eq:inf}
  \risk_P(H)
  \le
  \inf_{D\in\CCC}
  \risk_P(D)
  +
  \epsilon
\end{equation}
with probability at least $1-\delta$
when given $N$ examples generated independently
from any probability distribution $P$.
The \emph{sample complexity} of $\CCC$ can now be defined as the smallest $N$
with this property.
\begin{corollary}\label{cor:main2}
  For every $R>0$,
  the sample complexity of the \emph{restricted $R$-ball}
  \begin{equation*}
    \left\{
      D\in\FFF
      \st
      \left\|D\right\|_{\FFF}\le R
      \;\&\;
      D(\mathbf{X})\subseteq[-Y,Y]
    \right\}
  \end{equation*}
  in $\FFF$ does not exceed
  \begin{equation}\label{eq:cor2}
    \frac{8Y^2}{\epsilon^2}
    \left(
      \left(
        \ccc_{\FFF}^2 + 1
      \right)
      (R+Y)^2
      +
      8 Y^2
      \ln\frac{2}{\delta}
    \right)
    +
    1.
  \end{equation}
\end{corollary}
\begin{proof}
  According to (\ref{eq:cor1}) we can take
  \begin{equation*}
    \epsilon
    =
    \frac{2Y}{\sqrt{N}}
    \left(
      \sqrt{\ccc_{\FFF}^2 + 1}
      \left(
        R + Y
      \right)
      +
      2Y
      \sqrt{2\ln\frac{2}{\delta}}
    \right).
  \end{equation*}
  Expressing $\surd N$ through $\epsilon$,
  we obtain
  \begin{multline*}
    N
    \le
    \left(
      \frac{2Y}{\epsilon}
      \left(
        \sqrt{\ccc_{\FFF}^2 + 1}
        \left(
          R + Y
        \right)
        +
        2Y
        \sqrt{2\ln\frac{2}{\delta}}
      \right)
    \right)^2
    +
    1\\
    \le
    \frac{8Y^2}{\epsilon^2}
    \left(
      \left(
        \ccc_{\FFF}^2 + 1
      \right)
      \left(
        R + Y
      \right)^2
      +
      8Y^2
      \ln\frac{2}{\delta}
    \right)
    +
    1
  \end{multline*}
  (we have used the inequality $(a+b)^2\le2a^2+2b^2$).
  \qedtext
\end{proof}
Notice, however, that Corollary \ref{cor:main2} is a statement
about a non-universal benchmark class;
therefore, it appears less interesting than Corollary \ref{cor:main1}.
\blueend\fi

\subsection*{Universally consistent procedures}

Suppose the object space $\mathbf{X}$
is the Euclidean space $\bbbr^m$ for some $m$.
It is easy to see that Corollary \ref{cor:main1}
implies the existence of universally consistent procedures
in the sense of Stone \cite{stone:1977}
for a known upper bound $Y$ on $\lvert y_n\rvert$.
Indeed, by Luzin's theorem
(\cite{dudley:2002}, Theorem 7.5.2 on p.~244;
see also Theorem 7.1.3 on p.~225)
for any Borel measurable prediction rule $f:\mathbf{X}\to[-Y,Y]$
and any $\epsilon>0$ there exist a closed set $F\subseteq\mathbf{X}$
of probability at least $1-\epsilon$
such that the restriction of $f$ to $A$ is continuous;
it is obvious
\ifFULL\bluebegin
  and also follows from Ulam's theorem
  (\cite{dudley:2002}, Theorem 7.1.4 on p.~225)
\blueend\fi
that we can also assume that $F$ is compact.
Let $D$ be a function in a universal RKHS on $\mathbf{X}$
(the existence of the latter is shown in \S\ref{sec:examples})
taking values in $[-Y,Y]$ and close to $f$ in the metric $C(F)$.
It remains to apply Corollary \ref{cor:main1}.

\ifFULL\bluebegin
Perhaps this derivation can also be done
in the case where the object space $\mathbf{X}$ is a metric space.
By Luzin's theorem
(\cite{dudley:2002}, Theorem 7.5.2 on p.~244;
see also Theorem 7.1.3 on p.~225)
for any Borel measurable prediction rule $f:\mathbf{X}\to[-Y,Y]$
and any $\epsilon>0$ there exists a closed set $F\subseteq\mathbf{X}$
of probability at least $1-\epsilon$
such that the restriction of $f$ to $A$ is continuous;
by Ulam's theorem
(\cite{dudley:2002}, Theorem 7.1.4 on p.~225)
we can also assume that $F$ is compact.
We need a universal RKHS on $\mathbf{X}$,
or maybe the idea in the remark on p.~\pageref{p:remark2}
will work.
\blueend\fi

Intuitively,
the statistical assumption that the examples are produced
independently from the same distribution is strong enough
for the requirement of continuity to be superfluous:
as Cover mentioned in his discussion of Stone's paper,
it holds automatically with high probability.

\section{Examples of RKHS\ifnotCONF{} and reproducing kernels\fi}
\label{sec:examples}

The usefulness of the results stated in the previous two sections
depends on the availability of suitable RKHS.
In this section I will only give simplest examples;
for numerous other examples see, e.g.,
\cite{vapnik:1998}, \cite{scholkopf/smola:2002}, and \cite{shawe-taylor/cristianini:2004}.

\ifnotCONF\subsection*{The Sobolev spaces}\fi

The \emph{Sobolev norm} $\left\|f\right\|_{H^1}$
of an absolutely continuous function $f:[0,1]\to\bbbr$
is defined by
\begin{equation}\label{eq:norm-sobolev}
  \left\|f\right\|_{H^1}^2
  :=
  \int_0^1
    \left(
      f(t)
    \right)^2
  \D t
  +
  \int_0^1
    \left(
      f'(t)
    \right)^2
  \D t.
\end{equation}
The \emph{Sobolev space} $H^1([0,1])$ on $[0,1]$
is the set of absolutely continuous $f:[0,1]\to\bbbr$
satisfying
$\left\|f\right\|_{H^1} < \infty$
equipped with the norm $\left\|\cdot\right\|_{H^1}$.
It is easy to see that $H^1([0,1])$ is an RKHS.

In fact, $H^1([0,1])$ is only one of a range of Sobolev spaces;
see, e.g., \cite{adams/fournier:2003}
for the definition of the full range
(denoted $W^{s,p}(\Omega)$ there;
we are interested in the case $s=1$, $p=2$, and $\Omega=(0,1)$,
with the elements of $W^{1,2}((0,1))$ extended to $[0,1]$ by continuity).
The space $H^1([0,1])$ is the ``least smooth''
among the Sobolev spaces $H^s([0,1])$
if we ignore the slightly less natural case of a fractional~$s$.
All of $H^s([0,1])$ are universal RKHS,
but $H^1([0,1])$ is a proper superset of all other $H^s([0,1])$,
and so is the ``most universal'' Sobolev space of this type.


\ifnotCONF
It is easy to see that neither of the two addends in (\ref{eq:norm-sobolev})
can be omitted:
if the first addend is omitted,
the square root of the right-hand side of (\ref{eq:norm-sobolev})
ceases to be a norm
(since it becomes zero for every constant),
and if the second addend is omitted,
the function space ceases to be an RKHS
(since the evaluation functionals become unbounded).
We can, however, ``partially omit'' the first addend
replacing (\ref{eq:norm-sobolev})
with the \emph{Fermi--Sobolev norm}
$\left\|f\right\|_{\FS}$
defined by
\begin{equation}\label{eq:norm-fermi-sobolev}
  \left\|f\right\|_{\FS}^2
  :=
  \left(
    \int_0^1
      f(t)
    \D t
  \right)^2
  +
  \int_0^1
    \left(
      f'(t)
    \right)^2
  \D t
\end{equation}
for absolutely continuous functions $f:[0,1]\to\bbbr$.
The \emph{Fermi--Sobolev space on $[0,1]$}
is the set of absolutely continuous $f:[0,1]\to\bbbr$
satisfying
$\left\|f\right\|_{\FS} < \infty$
equipped with the norm $\left\|\cdot\right\|_{\FS}$.
It is clear that it is still an RKHS,
and it is still universal.

\ifFULL\bluebegin
\begin{remark*}
  The Fermi--Sobolev space is very popular in statistics:
  see, e.g., \cite{wahba:1990}, pp.~129--130,
  or \cite{gu:2002}, \S2.3.3.
  Surprisingly, it does not have a name;
  in this draft I use the name I gave to it
  in the early versions of \cite{\GTPXIII},
  before learning how popular it was.
\end{remark*}
\blueend\fi

Of course, the underlying set $Z$ of an RKHS
does not have to be a compact topological space:
we can define the Sobolev norm $\left\|f\right\|_{H^1}$
of an absolutely continuous function $f:\bbbr\to\bbbr$
by essentially the same formula
\begin{equation}\label{eq:norm-sobolev-infty}
  \left\|f\right\|_{H^1}^2
  :=
  \int_{-\infty}^{\infty}
    \left(
      f(t)
    \right)^2
  \D t
  +
  \int_{-\infty}^{\infty}
    \left(
      f'(t)
    \right)^2
  \D t
\end{equation}
and define the Sobolev space $H^1(\bbbr)$ on $\bbbr$
as the set of absolutely continuous $f:\bbbr\to\bbbr$
satisfying $\left\|f\right\|_{H^1} < \infty$.
\fi

To apply Theorems \ref{thm:main1}--\ref{thm:main3}
to \ifCONF $H^1([0,1])$\fi\ifnotCONF these RKHS\fi{}
we need to know the value of $\ccc_{\FFF}$ for
\ifCONF
  it; it is, however, well known (see \cite{\GTPXI} for details) that
  \begin{equation*}
    \ccc_{H^1([0,1])}
    =
    \sqrt{\coth 1}
    \approx
    1.15.
  \end{equation*}
\fi
\ifnotCONF
them; later in this section we will see that
\begin{equation*}
  \ccc_{\FFF}
  =
  \ccc_{H^1([0,1])}
  =
  \sqrt{\coth 1}
  \approx
  1.15
\end{equation*}
for the Sobolev space $H^1([0,1])$,
\begin{equation*}
  \ccc_{\FFF}
  =
  \ccc_{\FS}
  =
  2/\sqrt{3}
  \approx
  1.15
\end{equation*}
for the Fermi--Sobolev space on $[0,1]$,
and
\begin{equation*}
  \ccc_{\FFF}
  =
  \ccc_{H^1(\bbbr)}
  =
  1/\sqrt{2}
  \approx
  0.71
\end{equation*}
for the Sobolev space $H^1(\bbbr)$.

\begin{remark*}
  The term ``Sobolev space'' usually serves as the name for a topological vector space;
  all these spaces are normable,
  but different norms are not considered to lead
  to different Sobolev spaces
  as long as the topology does not change.
  The norms given by (\ref{eq:norm-sobolev}) and (\ref{eq:norm-sobolev-infty})
  are the most standard ones.
  It is easy to see that the norm (\ref{eq:norm-fermi-sobolev})
  leads to the same topology as (\ref{eq:norm-sobolev}):
  $\left\|f\right\|_{\FS}\le\left\|f\right\|_{H^1}$
  follows from the standard inequality between the $L^1$ and $L^2$ norms,
  and
  $\left\|f\right\|_{H^1}=O(\left\|f\right\|_{\FS})$
  follows from Wirtinger's inequality
  (which implies that $\int_0^{\pi} f^2 \le \int_0^{\pi} (f')^2$
  for every function $f$ on $[0,\pi]$ such that $f'$ is in $L_2$
  and $\int_0^{\pi} f = 0$;
  for the statement and a proof of Wirtinger's inequality,
  see \cite{hardy/etal:1952}, Theorem~258).
\end {remark*}
\fi

We are often interested in the case where the objects $x_n$ are vectors
in a Euclidean space $\bbbr^m$;
if their components are bounded,
we can scale them so that $x_n\in[0,1]^m$.
In any case,
we can take the $m$th tensor power $\FFF$
\ifnotCONF
 of one of the three RKHS
\fi
\ifCONF
 of the RKHS
\fi
we have just defined
as our benchmark class.
(For the definition and properties of tensor products of RKHS
see, e.g., \cite{aronszajn:1950}, \S I.8.)
\ifnotCONF
  We will see later that
\fi
\ifCONF
  The value of
\fi
$\ccc_{\FFF}$ for the $m$th tensor power
is the $m$th power of the $\ccc_{\FFF}$ for the original RKHS.
The $m$th tensor power
\ifCONF
  of $H^1([0,1])$ is universal on $[0,1]^m$
\fi
\ifnotCONF
  of the Sobolev and Fermi--Sobolev spaces on $[0,1]$ are universal on $[0,1]^m$
  and the $m$th tensor power of $H^1(\bbbr)$ is universal on $\bbbr^m$
\fi
(this can be seen
from the construction given in \cite{aronszajn:1950}, \S I.8).

Theorem \ref{thm:main3} requires separable RKHS;
the separability of Sobolev spaces $H^s$ for integer $s$
is proved in, e.g., \cite{adams/fournier:2003}, Theorem 3.6
(and it also remains true for fractional $s$).

\ifnotCONF
\subsection*{Reproducing kernels}

An equivalent language for talking about RKHS
is provided by the notion of a reproducing kernel;
this subsection defines reproducing kernels
and summarizes some of their properties.
For a detailed discussion,
see, e.g., \cite{aronszajn:1944}--\cite{aronszajn:1950} or \cite{meschkowski:1962}.

Let $\FFF$ be an RKHS on $Z$.
By the Riesz--Fischer theorem,
for each $z\in Z$ there exists a function $\kkk_z\in\FFF$ such that
\begin{equation}\label{eq:reproducing}
  f(z)
  =
  \langle \kkk_z,f\rangle_{\FFF},
  \quad
  \forall f\in\FFF.
\end{equation}
The next lemma asserts that $\left\|\kkk_z\right\|_{\FFF}$
is the norm $\ccc_{\FFF}(z)$ of the evaluation functional $f\mapsto f(z)$.
\begin{lemma}\label{lem:optimization}
  Let $\FFF$ be an RKHS on $Z$.
  For each $z\in Z$,
  \begin{equation*}
    \left\|\kkk_z\right\|_{\FFF}
    =
    \ccc_{\FFF}(z).
  \end{equation*}
\end{lemma}
\begin{proof}
  Fix $z\in Z$.
  We are required to prove
  \begin{equation*}
    \sup_{f:\left\|f\right\|_{\FFF}\le1}
    \left|
      f(z)
    \right|
    =
    \left\|\kkk_z\right\|_{\FFF}.
  \end{equation*}
  The inequality $\le$ follows from
  \begin{equation*}
    \left|
      f(z)
    \right|
    =
    \left|
      \left\langle
        f,\kkk_z
      \right\rangle_{\FFF}
    \right|
    \le
    \left\|
      f
    \right\|_{\FFF}
    \left\|
      \kkk_z
    \right\|_{\FFF}
    \le
    \left\|
      \kkk_z
    \right\|_{\FFF},
  \end{equation*}
  where $\left\|f\right\|_{\FFF}\le1$.
  The inequality $\ge$ follows from
  \begin{equation*}
    \left|
      f(z)
    \right|
    =
    \frac
    {\kkk_z(z)}
    {
      \left\|
        \kkk_z
      \right\|_{\FFF}
    }
    =
    \frac
    {
      \left\langle
        \kkk_z,\kkk_z
      \right\rangle_{\FFF}
    }
    {
      \left\|
        \kkk_z
      \right\|_{\FFF}
    }
    =
    \left\|
      \kkk_z
    \right\|_{\FFF},
  \end{equation*}
  where $f:=\kkk_z/\left\|\kkk_z\right\|_{\FFF}$
  and $\left\|\kkk_z\right\|_{\FFF}$ is assumed to be non-zero.
  \qedtext
\end{proof}

The \emph{reproducing kernel} of $\FFF$ is the function $\kkk:Z^2\to\bbbr$ defined by
\begin{equation*}
  \kkk(z,z')
  :=
  \left\langle
    \kkk_z,\kkk_{z'}
  \right\rangle_{\FFF}
\end{equation*}
(equivalently, we could define $\kkk(z,z')$ as $\kkk_z(z')$
or as $\kkk_{z'}(z)$).
The origin of this name is the ``reproducing property'' (\ref{eq:reproducing}).

There is a simple internal characterization of reproducing kernels of RKHS.
First,
it is easy to check that the function $\kkk(z,z')$,
as we defined it,
is symmetric,
\begin{equation*}
  \kkk(z,z')=\kkk(z',z),
  \quad
  \forall (z,z')\in Z^2,
\end{equation*}
and positive definite,
\begin{multline*}
  \sum_{i=1}^m\sum_{j=1}^m \alpha_i\alpha_j \kkk(z_i,z_j)\ge0,\\
  \forall m=1,2,\ldots,
  (\alpha_1,\ldots,\alpha_m)\in\bbbr^m,
  (z_1,\dots,z_m)\in Z^m.
\end{multline*}
On the other hand,
for every symmetric and positive definite $\kkk:Z^2\to\bbbr$
there exists a unique RKHS $\FFF$
such that $\kkk$ is the reproducing kernel of $\FFF$
(\cite{aronszajn:1944}, Theorem 2 on p.~143).

We can see that the notions of a reproducing kernel of RKHS
and of a symmetric positive definite function on $Z^2$
have the same content,
and we will sometimes say ``kernel on $Z$''
to mean a symmetric positive definite function
on $Z^2$.
Kernels in this sense are the main source of RKHS in learning theory:
cf.\ \cite{vapnik:1998,scholkopf/smola:2002,shawe-taylor/cristianini:2004}.
Every kernel on $\mathbf{X}$ is a valid parameter
for our prediction algorithms;
to apply Theorems \ref{thm:main1}--\ref{thm:main3}
we can use the equivalent definition
of $\ccc_{\FFF}$,
\begin{equation}\label{eq:equiv}
  \ccc_{\FFF}
  =
  \ccc_{\kkk}
  :=
  \sup_{x\in\mathbf{X}}
  \sqrt{\kkk(x,x)},
\end{equation}
$\kkk$ being the reproducing kernel of $\FFF$.

It was convenient to start from RKHS in stating the theorems
of \S\ref{sec:main},
but our prediction algorithms,
two of which are explicitly described in \S\ref{sec:algorithms},
use the more constructive representation of RKHS
via their reproducing kernels.

\ifFULL\bluebegin
\subsection*{Reproducing kernels II}

A popular third way is to start from a picture
that involves both notions,
closely intertwined.
A function $\kkk:\mathbf{X}^2\to\bbbr$
is said to be a \emph{reproducing kernel II}
of a Hilbert space $\FFF$ of functions on $\mathbf{X}$
if:
\begin{itemize}
\item
  for every $x\in\mathbf{X}$,
  \begin{equation}\label{eq:reproducing1}
    \kkk(\cdot,x)
    \in
    \FFF;
  \end{equation}
\item
  for all $f\in\FFF$ and $x\in\mathbf{X}$,
  \begin{equation}\label{eq:reproducing2}
    f(x)
    =
    \langle
      f(\cdot),\kkk(\cdot,x)
    \rangle_{\FFF}.
  \end{equation}
\end{itemize}
Since a Hilbert function space
can never have more than one reproducing kernel II
(\cite{aronszajn:1950}, \S I.2, (1)),
we will say that such a $\kkk$
is \emph{the} reproducing kernel II of $\FFF$.

All three ways are equivalent
in the following sense:
\begin{itemize}
\item
  if $\FFF$ is an RKHS on $\mathbf{X}$,
  its reproducing kernel $\kkk$ is a kernel on $\mathbf{X}$
  satisfying (\ref{eq:reproducing1}) and (\ref{eq:reproducing2});
\item
  if $\FFF$ is a Hilbert space $\FFF$ of functions on $\mathbf{X}$
  with reproducing kernel II $\kkk$,
  then $\kkk$ is a kernel on $\mathbf{X}$
  (\cite{aronszajn:1944}, \S I.2, (2,2) and (2,4))
  and $\FFF$ is an RKHS
  (\cite{aronszajn:1944}, Th\'eor\`eme 1)
  with reproducing kernel $\kkk$
  (this follows immediately from (\ref{eq:reproducing2}));
\item
  if $\kkk$ is a kernel on $\mathbf{X}$,
  there is one and only one Hilbert function space $\FFF$
  with $\kkk$ as its reproducing kernel II
  (\cite{aronszajn:1944}, Th\'eor\`eme 2)
  and
  there is one and only one RKHS $\FFF$
  with $\kkk$ as its reproducing kernel
  (this follows from the previous statements).
\end{itemize}

\blueend\fi

\subsection*{Norm vs.\ the reproducing kernel in RKHS}

Finding the norm given the reproducing kernel and vice versa
are often nontrivial problems for specific RKHS.
The most popular methods appear to be the following.
\begin{itemize}
\item
  As we saw in the proof of Lemma \ref{lem:optimization},
  $\kkk_z/\left\|\kkk_z\right\|_{\FFF}$
  is the function at which
  \begin{equation*}
    \sup_{f:\left\|f\right\|_{\FFF}\le1}
    \left|
      f(z)
    \right|
  \end{equation*}
  is attained
  (assuming that $\left\|\kkk_z\right\|_{\FFF}\ne0$
  and that this optimization problem has a unique solution).
  Solving this optimization problem we can find the kernel $\kkk$
  given the norm $f\mapsto\left\|f\right\|_{\FFF}$.
  For application of this method to the Fermi--Sobolev space on $[0,1]$,
  see \cite{\GTPXIII}, Appendix C\ifnotarXiv{ of the last arXiv version}\fi.
\item
  One can use expansions into Fourier series of functions in a given RKHS.
  For examples see, e.g.,
  \cite{gu:2002}, \S4.2.1,
  or, for the Fermi--Sobolev space on $[0,1]$,
  \cite{\GTPXIII} (version 2\ifnotarXiv\ of the arXiv technical report\fi).
\item
  If $Z$ is a Euclidean space
  and the reproducing kernel $\kkk(z,z')$ only depends on the difference $z-z'$
  (is ``translation-invariant''),
  an explicit formula for the reproducing kernel
  can sometimes be obtained by applying the Fourier transform
  to both sides of (\ref{eq:reproducing})
  (similar methods are applied to the Sobolev space $H^1(\bbbr)$
  in \cite{thomas-agnan:1996} and \cite{smola/etal:1998}).
\end{itemize}

The reproducing kernel of the Sobolev space $H^1([0,1])$,
as given in \cite{berlinet/thomas-agnan:2004}
(\S7.4, Example 13; Exercise 3.12.7)
with a reference to \cite{atteia:1992},
is
\begin{equation*}
  \kkk(t,t')
  =
  \frac
  {
    \cosh
    \min(t,t')
    \cosh
    \min(1-t,1-t')
  }
  {
    \sinh 1
  }.
\end{equation*}
This implies Marti's \cite{marti:1983} result that
\begin{equation*}
  \ccc^2_{\kkk}
  =
  \sup_{t\in[0,1]}
  \frac
  {
    \cosh t
    \cosh(1-t)
  }
  {
    \sinh 1
  }
  =
  \frac
  {
    \cosh 0
    \cosh 1
  }
  {
    \sinh 1
  }
  =
  \coth 1,
\end{equation*}
as stated above.

The reproducing kernel of the Fermi--Sobolev space on $[0,1]$
was found in \cite{craven/wahba:1979}
(see also \cite{wahba:1990}, \S10.2, or \cite{gu:2002}, \S2.3.3);
it is given by
\begin{align}
  \kkk(t,t')
  &=
  k_0(t) k_0(t')
  +
  k_1(t) k_1(t')
  +
  k_2(\lvert t-t'\rvert)\notag\\
  &=
  1
  +
  \left(
    t-\frac12
  \right)
  \left(
    t'-\frac12
  \right)
  +
  \frac12
  \left(
    \lvert t-t' \rvert^2
    -
    \lvert t-t' \rvert
    +
    \frac16
  \right)\notag\\
  &=
  \frac12
  \minsq(t,t')
  +
  \frac12
  \minsq(1-t,1-t')
  +
  \frac56,
  \label{eq:final}
\end{align}
where $k_l:=B_l/l!$ are scaled Bernoulli polynomials $B_l$.
So, for the Fermi--Sobolev space on $[0,1]$ we have
\begin{equation*}
  \ccc^2_{\kkk}
  =
  \max_{t\in[0,1]}
  \left(
    \frac12
    t^2
    +
    \frac12
    (1-t)^2
    +
    \frac56
  \right)
  =
  \frac43.
\end{equation*}

The reproducing kernel of the Sobolev space $H^1(\bbbr)$ is
\begin{equation*}
  \kkk(t,t')
  =
  \frac{1}{2}
  \exp
  \left(
    -
    \left|
      t-t'
    \right|
  \right)
\end{equation*}
(see \cite{thomas-agnan:1996},
\cite{smola/etal:1998},
or \cite{berlinet/thomas-agnan:2004}, \S7.4, Example 24).
From the last equation we can see that $\ccc_{H^1(\bbbr)}=1/\sqrt{2}$.

It is the general fact that the reproducing kernel of the $m$-fold product of RKHS
can be obtained as the $m$-fold product of the reproducing kernels
of the components
(\cite{aronszajn:1950}, \S I.8, Theorem~I).
For example,
the reproducing kernel of the $m$th power of $H^1([0,1])$ is
\begin{equation*}
  \kkk
  \left(
    (t_1,\ldots,t_m),
    (t'_1,\ldots,t'_m)
  \right)
  =
  \prod_{i=1}^m
  \frac
  {
    \cosh
    \min(t_i,t'_i)
    \cosh
    \min(1-t_i,1-t'_i)
  }
  {
    \sinh 1
  }.
\end{equation*}
We can see that
\begin{equation*}
  \ccc_{\FFF}
  =
  \left(
    \coth 1
  \right)^{m/2},
  \quad
  \ccc_{\FFF}
  =
  \left(
    2/\sqrt{3}
  \right)^m,
  \quad
  \ccc_{\FFF}
  =
  2^{-m/2}
\end{equation*}
for the $m$th power of the Sobolev space $H^1([0,1])$,
of the Fermi--Sobolev space on $[0,1]$,
and of the Sobolev space $H^1(\bbbr)$,
respectively.

An extensive list of RKHS together with their reproducing kernels is given
in \cite{berlinet/thomas-agnan:2004}, \S7.4.
\fi

\section{Some comparisons}
\label{sec:review}

\ifFULL\bluebegin
In this section I will state several results from literature
related to Theorems \ref{thm:main1} and \ref{thm:main2}
(in the first subsection)
and Corollaries \ref{cor:main1}--\ref{cor:main3}
(in the second subsection)
that I happen to know about.
This part of the paper is especially sketchy.

\subsection*{Competitive on-line learning}
\blueend\fi

The first paper about competitive on-line regression is \cite{foster:1991};
for a brief review of the work done in the 1990s,
see \cite{vovk:2001competitive}, \S4.
Our results are especially close to those of
\cite{cesabianchi/long/warmuth:1996}
and \cite{auer/etal:2002}.

There are two main proof techniques
in the existing theory of competitive on-line regression:
various generalizations of gradient descent
(used in, e.g.,
\cite{cesabianchi/long/warmuth:1996},
\cite{kivinen/warmuth:1997},
and \cite{auer/etal:2002})
and the Bayes-type Aggregating Algorithm
(proposed in \cite{vovk:1990} and described in detail
in \cite{haussler/etal:1998};
for a streamlined presentation, see \cite{vovk:2001competitive}).
In this subsection we will only discuss the former;
some information about the latter will be given in \S\ref{sec:proof3}.

Comparison between our results and the known ones
is somewhat complicated by the fact
that most of the existing literature
only deals with the Euclidean spaces $\bbbr^m$.
Typically, when loss bounds do not depend on $m$,
they can be carried over to Hilbert spaces
(perhaps satisfying some extra regularity assumptions,
such as separability),
and so to some RKHS.
To understand what such known results
say in the case of RKHS,
the upper bound on the size $\left\|x_n\right\|$ of the objects
(if present) has to be replaced by $\ccc_{\FFF}$
(cf.\ the remark on p.~\pageref{p:remark1}),
and the upper bound on the size $\left\|w\right\|$ of the weight vector
has to be interpreted as an upper bound on $\left\|D\right\|_{\FFF}$.

With such replacements,
Theorem IV.4 on p.~610 of Cesa-Bianchi \emph{et al.}\ \cite{cesabianchi/long/warmuth:1996}
becomes
\begin{multline*}
  \sum_{n=1}^N
  (y_n-\mu_n)^2
  \le
  \inf_{D:\left\|D\right\|_{\FFF}\le Y/X}
  \sum_{n=1}^N
  (y_n-D(x_n))^2\\
  +
  9.2
  \left(
    Y
    \sqrt
    {
      \inf_{D:\left\|D\right\|_{\FFF}\le Y/X}
      \sum_{n=1}^N
      (y_n-D(x_n))^2
    }
    +
    Y^2
  \right),
\end{multline*}
where $\mu_n$ are their algorithm's predictions.
This result is of the same type as (\ref{eq:goal2}),
but $\left\|D\right\|_{\FFF}$ is bounded by $Y/X$;
because of such a bound
(present in all other results reviewed here)
the corresponding prediction algorithm
is not guaranteed to be universally consistent.

Auer \emph{et al.}\ \cite{auer/etal:2002}
make the upper bound on $\left\|D\right\|_{\FFF}$ more general:
their Theorem 3.1 (p.~66) implies that,
for their algorithm,
\begin{multline*}
  \sum_{n=1}^N
  (y_n-\mu_n)^2
  \le
  \sum_{n=1}^N
  (y_n-D(x_n))^2\\
  +
  8
  \ccc_{\FFF}^2
  U^2
  +
  8
  \ccc_{\FFF}
  U
  \sqrt
  {
    \frac12
    \sum_{n=1}^N
    (y_n-D(x_n))^2
    +
    \ccc_{\FFF}^2
    U^2
  },
\end{multline*}
where $U$ is a known upper bound on $\left\|D\right\|_{\FFF}$
and $Y$ is assumed to be $1$.
This is remarkably similar to (\ref{eq:goal2}) and (\ref{eq:goal3}).

This type of results was extended by Zinkevich
(\cite{zinkevich:2003}, Theorem 1)
to a general class of convex loss functions.

The main differences of these results from our Theorems \ref{thm:main1}--\ref{thm:main3}
are that their leading constants are somewhat worse
and that they assume a known upper bound on $\left\|D\right\|_{\FFF}$.
The last circumstance might appear especially serious,
since it prevents universal consistency
even when the Hilbert space used is a universal RKHS.
However,
there is a simple way to achieve universal consistency:
the Aggregating Algorithm, or a similar procedure,
may be used on top of the existing algorithm
(the unknown upper bound may be considered to be an ``expert'',
and the predictions made by all ``experts'', say of the form $2^k$, $k=1,2,\ldots$,
can be merged into one prediction on each round).
This was noticed by Auer \emph{et al.}\ \cite{auer/etal:2002},
although they did not develop this idea further.

The remaining minor component in achieving universal consistency
is using a universal function class as the benchmark class.
It is interesting that Cesa-Bianchi \emph{et al.}\ used
an ``almost universal'' function class
in their pioneering paper \cite{cesabianchi/long/warmuth:1996}
(\S V;
their class was not quite universal because of the requirement $f(0)=0$).
A very interesting early paper about on-line regression
competitive with function spaces (although not universal)
is \cite{kimber/long:1995}
(continued by \cite{long:2000});
it, however, assumes that the benchmark class
contains a perfect prediction rule,
and its results are very different from ours.

A major advantage of the methods based on gradient descent
is their simplicity and computational efficiency.
The technique of defensive forecasting,
which we emphasize in this paper,
appears closer to gradient descent than to the Bayes-type algorithms.
There has been a mutually beneficial exchange of ideas between
the gradient descent and Bayes-type approaches,
and combining gradient descent and defensive forecasting
might turn out even more productive.

Results such as Corollary \ref{cor:main1} can be obtained by a routine application
of well-known
results in competitive on-line learning,
but they might not be easy to obtain by the traditional methods
of statistical learning theory.
The closest results of this kind in statistical learning theory
that I am aware of are Theorem C${}^*$
(applied to Sobolev spaces and smooth kernels in Examples 3 and 4)
of \cite{cucker/smale:2002}
and Corollary 6.7 of \cite{bartlett/etal:2005}.
These results, however, use balls in RKHS as benchmark classes,
and therefore, do not guarantee even universal consistency.

Corollary \ref{cor:main1} can be strengthened by using
the results of \cite{cesabianchi/gentile:2005}
instead of those of \cite{cesabianchi/etal:2004}.

\ifFULL{\text{blue}
\subsection*{Statistical theory of regression}

In this subsection we will review results 
related to our Corollaries \ref{cor:main1}--\ref{cor:main3}.
Some of these results depend on extra assumptions about the RKHS considered,
but these assumptions are relatively mild and satisfied
\ifnotCONF
  in most examples
\fi
\ifCONF
  in the examples
\fi
considered in the previous section.
We will use the same notation that we used
in Corollaries \ref{cor:main1}--\ref{cor:main3}.

Cucker and Smale's paper \cite{cucker/smale:2002}
reports a result closest to ours.
Let $\mathbf{X}$ be a compact domain or manifold in $\bbbr^m$
and $H^s(\mathbf{X})$, $s>m/2$,
be the Sobolev space on it
(for the definition in the case $\mathbf{X}=[0,1]$ and $s=1$,
see the previous section).
Cucker and Smale (\cite{cucker/smale:2002}, Example 3 (continued) on p.~18)
obtain the upper bound
\begin{equation}\label{eq:cucker-smale-2}
  \frac{1152 Y^2}{\epsilon}
  \left(
    \left(
      \frac{48CRY}{\epsilon}
    \right)^{m/s}
    +
    \ln\frac{1}{\delta}
  \right),
\end{equation}
where $C$ is a constant
(which may, however, depend on $s$),
on the sample complexity of the restricted $R$-ball in $H^s(\mathbf{X})$.
Therefore, for the space $H^1([0,1])$ we have two bounds on the sample complexity:
\begin{align*}
  &\frac{8Y^2}{\epsilon^2}
  \left(
    \left(
      \coth 1 + 1
    \right)
    (R+Y)^2
    +
    8 Y^2
    \ln\frac{2}{\delta}
  \right)
  +
  1
  &&\text{Corollary \ref{cor:main2}}\\
  &\frac{1152 Y^2}{\epsilon}
  \left(
    \frac{48CRY}{\epsilon}
    +
    \ln\frac{1}{\delta}
  \right)
  &&\text{(\ref{eq:cucker-smale-2}) with $s=1$}.
\end{align*}
Both bounds scale as $\epsilon^{-2}$ for small $\epsilon>0$.
Neither is strictly better than the other:
e.g.,
an advantage of our bound is that all constants are given explicitly and are rather small,
and an advantage of Cucker and Smale's bound
is that the coefficient in front of $\ln(1/\delta)$
scales as $\epsilon^{-1}$.
Going outside $H^1([0,1])$,
we have a similar bound, (\ref{eq:cor2}), for general RKHS
(e.g., we do not need the compactness of $\mathbf{X}$ for Sobolev spaces);
on the other hand,
Cucker and Smale's bounds scale better in $\epsilon$
for classes of smooth functions
(cf.\ (\ref{eq:cucker-smale-2}) for $s>1$ and, especially,
the result of \cite{cucker/smale:2002}, Example 4 (continued) on p.~18,
for $C^{\infty}$ kernels;
it should be noted, however, that our bound scales better in $\epsilon$
for the least smooth of the Sobolev classes $H^s([0,1])$ that are RKHS,
namely,
for $s\in(1/2,1)$).

Cucker and Smale's methods are based on using entropy numbers of compact operators
(introduced into learning theory by
\cite{williamson/etal:2001}).
A very different method, based on stability of learning algorithms,
is proposed by Bousquet and Elisseeff,
who prove the bound
\begin{equation*}
  \risk_P(H_N)
  \le
  \frac1N
  \sum_{n=1}^N
  \bigl(
    y_n - H_N(x_n)
  \bigr)^2
  +
  \frac{16 Y^2 \ccc_{\FFF}^2}{\lambda N}
  +
  \left(
    \frac{32 Y^2 \ccc_{\FFF}^2}{\lambda}
    +
    4Y
  \right)
  \sqrt{\frac{\ln\frac{1}{\delta}}{2N}}
\end{equation*}
for the prediction rule $H_N=H$
obtained by the regularized least squares,
\begin{equation*}
  \frac1N
  \sum_{n=1}^N
  \bigl(
    y_n - H(x_n)
  \bigr)^2
  +
  \lambda
  \left\|
    H
  \right\|_{\FFF}^2
  \to
  \min,
\end{equation*}
where $\lambda$ is a positive constant
(\cite{elisseeff/bousquet:2002},
Example 3 in \S5.2.2, p.~517).
Therefore, we have, with probability $1-\delta$ for all $D$ simultaneously,
\begin{multline}\label{eq:be}
  \risk_P(H_N)
  \le
  \frac1N
  \sum_{n=1}^N
  \bigl(
    y_n - D(x_n)
  \bigr)^2\\
  +
  \lambda
  \left\|
    D
  \right\|_{\FFF}^2
  +
  \frac{16 Y^2 \ccc_{\FFF}^2}{\lambda N}
  +
  \left(
    \frac{32 Y^2 \ccc_{\FFF}^2}{\lambda}
    +
    4Y
  \right)
  \sqrt{\frac{\ln\frac{1}{\delta}}{2N}}.
\end{multline}
This is similar to our (\ref{eq:cor3}),
except that the part following the first ``$+$'' in (\ref{eq:be})
does not tend to 0 for a fixed $\lambda$.
Taking $\lambda=\lambda_N\to0$ makes the dependence on $N$
in (\ref{eq:be}) worse than in (\ref{eq:cor3}).

There are several other results of this kind in literature,
such as
\ifFULL\bluebegin
\begin{equation*}
  \sqrt{\risk_P(H_N)}
  \le
  \sqrt
  {
    \risk_P(D)
    +
    \lambda
    \left\|
      D
    \right\|_{\FFF}^2
  }
  +
  \frac{Y \ccc_{\FFF}^2}{\lambda \sqrt{N}}
  +
  \left(
    1 + \frac{\ccc_{\FFF}}{\sqrt{\lambda}}
  \right)
  \left(
    1 + \sqrt{2\ln\frac{2}{\delta}}
  \right)
\end{equation*}
(\cite{devito/etal:2005}, Theorem 1)
and\blueend\fi
\begin{equation*}
  \Expect\risk_P(H_N)
  \le
  \left(
    1
    +
    \frac{2\ccc_{\FFF}^2}{\lambda N}
  \right)^2
  \left(
    \risk_P(D)
    +
    \frac{\lambda}{2}
    \left\|
      D
    \right\|_{\FFF}^2
  \right)
\end{equation*}
(\cite{zhang:2003}, as stated in \cite{smale/zhou:2005}, (7.17));
this should be compared to Corollary \ref{cor:main1}.
Some further information can be found in \cite{mukherjee/etal:2003}.

The traditional approach to regression in statistical learning theory
\cite{vapnik:1998}
is based on modifications of the VC dimension,
such as pseudo-dimension:
e.g., Lee, Bartlett, and Williamson \cite{lee/etal:1998}
prove that the sample complexity of a convex function class
of finite pseudo-dimension is
\begin{equation*}
  O
  \left(
    \frac{1}{\epsilon}
    \left(
      \ln\frac{1}{\epsilon}
      +
      \ln\frac{1}{\delta}
    \right)
  \right).
\end{equation*}
Unfortunately,
the pseudo-dimension is infinite for many interesting benchmark classes
(such as the examples of RKHS in \S\ref{sec:examples}).
\blueend\fi

\ifnotCONF
\section{Proof of Theorem~\ref{thm:main1}}
\label{sec:proof1}

This section is essentially a simplified (and to some degree cut-and-pasted)
version of \S\S 5--7 of \cite{\GTPXIV}.
First we modify the protocol of \S\ref{sec:main}
introducing a third player, Skeptic,
who is allowed to bet
at the odds implied by Predictor's moves.

\bigskip

\noindent
\textsc{Forecasting Game I}

\noindent
\textbf{Players:} Reality, Predictor, Skeptic

\noindent
\textbf{Protocol:}

\parshape=7
\IndentI   \WidthI
\IndentII  \WidthII
\IndentII  \WidthII
\IndentII  \WidthII
\IndentII  \WidthII
\IndentII  \WidthII
\IndentI   \WidthI
\noindent
FOR $n=1,2,\dots$:\\
  Reality announces $x_n\in\mathbf{X}$.\\
  Predictor announces $\mu_n\in\bbbr$.\\
  Skeptic announces $s_n\in\bbbr$.\\
  Reality announces $y_n\in[-Y,Y]$.\\
  $\K_n := \K_{n-1} + s_n (y_n-\mu_n)$.\\
END FOR.

\bigskip

\noindent
In this protocol,
the prediction $\mu_n$ is interpreted as the price
Predictor charges for a ticket paying $y_n$;
$s_n$ is the number of tickets Skeptic decides to buy.
(We sometimes refer to predictions interpreted this way
as forecasts,
although the difference between forecasts and the decision-type predictions
of \S\ref{sec:main} is not as important here
as for the more general loss functions considered in \cite{\GTPXIV}.)
The protocol describes not only the players' moves
but also the changes in Skeptic's capital $\K_n$;
its initial value $\K_0$ can be an arbitrary real number.
Protocols of this type are studied extensively in \cite{shafer/vovk:2001}.

For any continuous strategy for Skeptic
there exists a strategy for Predictor
that does not allow Skeptic's capital to grow,
regardless of Reality's moves.
To state this observation in its strongest form,
we make Skeptic announce his strategy
for each round
before Predictor's move on that round
rather than announce his full strategy at the beginning of the game.
Therefore,
we consider the following perfect-information game:

\bigskip

\noindent
\textsc{Forecasting Game II}

\noindent
\textbf{Players:} Reality, Predictor, Skeptic

\noindent
\textbf{Protocol:}

\parshape=7
\IndentI   \WidthI
\IndentII  \WidthII
\IndentII  \WidthII
\IndentII  \WidthII
\IndentII  \WidthII
\IndentII  \WidthII
\IndentI   \WidthI
\noindent
FOR $n=1,2,\dots$:\\
  Reality announces $x_n\in\mathbf{X}$.\\
  Skeptic announces continuous $S_n:\bbbr\to\bbbr$.\\
  Predictor announces $\mu_n\in\bbbr$.\\
  Reality announces $y_n\in[-Y,Y]$.\\
  $\K_n := \K_{n-1} + S_n(\mu_n) (y_n-\mu_n)$.\\
END FOR.

\bigskip

\begin{lemma}\label{lem:basic}
  Predictor has a strategy in Forecasting Game II
  that ensures $\mu_n\in[-Y,Y]$, for all $n=1,2,\ldots$,
  and $\K_0\ge\K_1\ge\K_2\ge\cdots$.
\end{lemma}
\begin{proof}
  Predictor's goal is achieved by the following strategy:
  \begin{itemize}
  \item
    if the function $S_n$ takes value 0 on the interval $[-Y,Y]$,
    choose $\mu_n\in[-Y,Y]$ such that $S_n(\mu_n)=0$;
  \item
    if $S_n$ is always positive on $[-Y,Y]$,
    take $\mu_n:=Y$;
  \item
    if $S_n$ is always negative on $[-Y,Y]$,
    take $\mu_n:=-Y$.
    \qedtext
  \end{itemize}
\end{proof}

\subsection*{Algorithm of Large Numbers}

We say that a kernel $\kkk$ on $[-Y,Y]\times\mathbf{X}$ is \emph{forecast-continuous}
if the function $\kkk((\mu,x),(\mu',x'))$ is continuous in $(\mu,\mu')\in[-Y,Y]^2$,
for all fixed $(x,x')\in\mathbf{X}^2$.
For such a kernel the function
\begin{equation}\label{eq:function1}
  S_n(\mu)
  :=
  \sum_{i=1}^{n-1}
  \kkk
  \bigl(
    (\mu,x_n),(\mu_i,x_i)
  \bigr)
  (y_i-\mu_i)
  -
  \kkk
  \bigl(
    (\mu,x_n),(\mu,x_n)
  \bigr)
  \mu
\end{equation}
is continuous in $\mu\in[-Y,Y]$.

\bigskip

\noindent
\textsc{The algorithm of large numbers (ALN)}

\noindent
\textbf{Parameter:} forecast-continuous kernel $\kkk$ on $[-Y,Y]\times\mathbf{X}$

\parshape=7
\IndentI   \WidthI
\IndentII  \WidthII
\IndentII  \WidthII
\IndentII  \WidthII
\IndentIII \WidthIII
\IndentII  \WidthII
\IndentI   \WidthI
\noindent
FOR $n=1,2,\dots$:\\
  Read $x_n\in\mathbf{X}$.\\
  Define $S_n:[-Y,Y]\to\bbbr$ by (\ref{eq:function1}).\\
  Output any root $\mu\in[-Y,Y]$ of $S_n(\mu)=0$ as $\mu_n$;\\
    if there are no roots, set $\mu_n:=Y\sign S_n$.\\
  Read $y_n\in[-Y,Y]$.\\
END FOR.

\bigskip

\noindent
(Notice that $\sign S_n$ is well defined in this context.)
It is well known that
for each kernel $\kkk$ on $[-Y,Y]\times\mathbf{X}$
there exists a function $\Phi:[-Y,Y]\times\mathbf{X}\to\HHH$
(a \emph{feature mapping} taking values in a Hilbert space $\HHH$)
such that
\begin{equation}\label{eq:K}
  \kkk(a,b)
  =
  \left\langle
    \Phi(a),\Phi(b)
  \right\rangle_{\HHH},
  \enspace
  \forall a,b\in[-Y,Y]\times\mathbf{X}.
\end{equation}
(For example,
we can take the RKHS on $[-Y,Y]\times\mathbf{X}$ with reproducing kernel $\kkk$ as $\HHH$
and take $a\mapsto\kkk_a$ as the feature mapping $\Phi$;
there are, however,
easier and more transparent constructions.)
It can be shown that $\Phi(\mu,x)$ is \emph{forecast-continuous},
i.e., continuous in $\mu\in[-Y,Y]$ for each fixed $x\in\mathbf{X}$,
if and only if the kernel $\kkk$ defined by~(\ref{eq:K}) is forecast-continuous
(see, e.g., \cite{\GTPXIII}, Appendix~B,
where $[0,1]$ should be replaced with $[-Y,Y]$).
\begin{theorem}\label{thm:ALN}
  Let $\kkk$ be the kernel
  defined by~(\ref{eq:K})
  for a forecast-continuous feature mapping $\Phi:[-Y,Y]\times\mathbf{X}\to\HHH$,
  where $\HHH$ is a Hilbert space.
  The ALN with parameter $\kkk$ outputs $\mu_n\in[-Y,Y]$
  such that
  \begin{equation}\label{eq:ALN}
    \left\|
      \sum_{n=1}^N
      (y_n-\mu_n)
      \Phi(\mu_n,x_n)
    \right\|^2_{\HHH}
    \le
    \sum_{n=1}^N
    \left(
      Y^2 - \mu_n^2
    \right)
    \left\|
      \Phi(\mu_n,x_n)
    \right\|_{\HHH}^2
  \end{equation}
  always holds for all $N=1,2,\dots$.
\end{theorem}
\begin{proof}
  Following the ALN,
  Predictor ensures that Skeptic will never increase his capital
  with the strategy
  \begin{equation}\label{eq:strategy}
    s_n
    :=
    \sum_{i=1}^{n-1}
    \kkk
    \bigl(
      (\mu_n,x_n),(\mu_i,x_i)
    \bigr)
    (y_i-\mu_i)
    -
    \kkk
    \bigl(
      (\mu_n,x_n),(\mu_n,x_n)
    \bigr)
    \mu_n.
  \end{equation}
  Using the inequalities
  \begin{equation*}
    (y_n-\mu_n)^2
    +
    2\mu_n
    \left(
      y_n - \mu_n
    \right)
    \le
    Y^2 - \mu_n^2
  \end{equation*}
  and
  \begin{equation*}
    \kkk
    \bigl(
      (\mu_n,x_n),(\mu_n,x_n)
    \bigr)
    \ge
    0
  \end{equation*}
  we can see that the increase in Skeptic's capital
  when he follows~(\ref{eq:strategy})
  is
  \begin{align*}
    \K_N-\K_0
    &=
    \sum_{n=1}^N
    s_n(y_n-\mu_n)\\
    &=
    \sum_{n=1}^N
    \sum_{i=1}^{n-1}
    \kkk
    \bigl(
      (\mu_n,x_n),(\mu_i,x_i)
    \bigr)
    (y_n-\mu_n)
    (y_i-\mu_i)\\
    &\quad{}-
    \sum_{n=1}^N
    \kkk
    \bigl(
      (\mu_n,x_n),(\mu_n,x_n)
    \bigr)
    \mu_n
    (y_n-\mu_n)\\
    &=
    \frac12
    \sum_{n=1}^N
    \sum_{i=1}^N
    \kkk
    \bigl(
      (\mu_n,x_n),(\mu_i,x_i)
    \bigr)
    (y_n-\mu_n)
    (y_i-\mu_i)\\
    &\quad{}-
    \frac12
    \sum_{n=1}^N
    \kkk
    \bigl(
      (\mu_n,x_n),(\mu_n,x_n)
    \bigr)
    (y_n-\mu_n)^2\\
    &\quad{}-
    \sum_{n=1}^N
    \kkk
    \bigl(
      (\mu_n,x_n),(\mu_n,x_n)
    \bigr)
    \mu_n
    (y_n-\mu_n)\\
    &\ge
    \frac12
    \sum_{n=1}^N
    \sum_{i=1}^N
    \kkk
    \bigl(
      (\mu_n,x_n),(\mu_i,x_i)
    \bigr)
    (y_n-\mu_n)
    (y_i-\mu_i)\\
    &\quad{}-
    \frac12
    \sum_{n=1}^N
    \kkk
    \bigl(
      (\mu_n,x_n),(\mu_n,x_n)
    \bigr)
    \left(
      Y^2 - \mu_n^2
    \right)\\
    &=
    \frac12
    \left\|
      \sum_{n=1}^N
      (y_n-\mu_n)
      \Phi(\mu_n,x_n)
    \right\|^2_{\HHH}
    -
    \frac12
    \sum_{n=1}^N
    \left(
      Y^2 - \mu_n^2
    \right)
    \left\|
      \Phi(\mu_n,x_n)
    \right\|^2_{\HHH},
  \end{align*}
  which immediately implies~(\ref{eq:ALN}).
  \qedtext
\end{proof}

\subsection*{Resolution}

This subsection makes the next step in our proof of Theorem~\ref{thm:main1}.
Our goal is to prove the following result
(although we will need a slight modification of this result
rather than the result itself).
\begin{theorem}\label{thm:resolution}
  Let $\FFF$ be an RKHS on $\mathbf{X}$
  with reproducing kernel $\kkk$.
  The forecasts $\mu_n\in[-Y,Y]$ output by the ALN with parameter $\kkk$ always satisfy
  \begin{equation*}
    \left|
      \sum_{n=1}^N
      (y_n-\mu_n)
      D(x_n)
    \right|
    \le
    Y
    \ccc_{\FFF}
    \left\|D\right\|_{\FFF}
    \sqrt{N}
  \end{equation*}
  for all $N$ and all functions $D\in\FFF$.
\end{theorem}
\begin{proof}
Using (\ref{eq:ALN}) with $\Phi$ being the feature mapping
$
  x\in\mathbf{X}
  \mapsto
  \kkk_x\in\FFF
$,
we obtain
\begin{multline}\label{eq:simple-resolution}
  \left|
    \sum_{n=1}^N
    (y_n-\mu_n)
    D(x_n)
  \right|
  =
  \left|
    \sum_{n=1}^N
    (y_n-\mu_n)
    \left\langle
      \kkk_{x_n}, D
    \right\rangle_{\FFF}
  \right|\\
  =
  \left|
    \left\langle
      \sum_{n=1}^N
      (y_n-\mu_n)
      \kkk_{x_n},
      D
    \right\rangle_{\FFF}
  \right|
  \le
  \left\|
    \sum_{n=1}^N
    (y_n-\mu_n)
    \kkk_{x_n}
  \right\|_{\FFF}
  \left\|
    D
  \right\|_{\FFF}\\
  \le
  \left\|
    D
  \right\|_{\FFF}
  \sqrt
  {
    \sum_{n=1}^N
    Y^2
    \kkk(x_n,x_n)
  }
  \le
  Y
  \ccc_{\FFF}
  \left\|
    D
  \right\|_{\FFF}
  \sqrt
  {
    N
  }
\end{multline}
for any $D\in\FFF$.
\qedtext
\end{proof}

Theorem \ref{thm:resolution} can be interpreted as asserting
that the ALN has a good ``resolution''
when $\FFF$ is a universal RKHS;
for details, see \cite{\GTPXIII}.

\subsection*{Mixing feature mappings}

In the proof of Theorem \ref{thm:main1}
we will mix the feature mapping $\Phi_0(\mu,x):=\mu$
(into $\HHH_0:=\bbbr$)
and the feature mapping $\Phi_1(\mu,x):=\kkk_x$
used in the proof of Theorem \ref{thm:resolution}
(we will have to achieve two goals simultaneously).
This can be done using the following corollary
of Theorem \ref{thm:ALN}.
\begin{corollary}\label{cor:mixture1}
  Let $\Phi_j:[-Y,Y]\times\mathbf{X}\to\HHH_j$, $j=0,1$,
  be forecast-continuous mappings from $[-Y,Y]\times\mathbf{X}$
  to Hilbert spaces $\HHH_j$,
  and let $a_0,a_1$ be two positive constants.
  The forecasts $\mu_n\in[-Y,Y]$ output by the ALN with a suitable kernel parameter always satisfy
  \begin{multline*}
    \left\|
      \sum_{n=1}^N
      (y_n-\mu_n)
      \Phi_j(\mu_n,x_n)
    \right\|^2_{\HHH_j}\\
    \le
    \frac{Y^2}{a_j}
    \sum_{n=1}^N
    \left(
      a_0
      \left\|
        \Phi_0(\mu_n,x_n)
      \right\|^2_{\HHH_0}
      +
      a_1
      \left\|
        \Phi_1(\mu_n,x_n)
      \right\|^2_{\HHH_1}
    \right)
  \end{multline*}
  for all $N$ and for both $j=0$ and $j=1$.
\end{corollary}
\begin{proof}
  Define the ``weighted direct sum'' $\HHH$ of $\HHH_0$ and $\HHH_1$
  as the Cartesian product $\HHH_0\times\HHH_1$
  equipped with the inner product
  \begin{equation*}
    \langle g, g' \rangle_{\HHH}
    =
    \left\langle
      (g_0,g_1),
      (g'_0,g'_1)
    \right\rangle_{\HHH}
    :=
    \sum_{j=0}^{1}
    a_j
    \langle g_j, g'_j\rangle_{\HHH_j}.
  \end{equation*}
  Now we can define $\Phi:[-Y,Y]\times\mathbf{X}\to\HHH$ by
  \begin{equation*}
    \Phi(\mu,x)
    :=
    \left(
      \Phi_0(\mu,x),
      \Phi_1(\mu,x)
    \right);
  \end{equation*}
  the corresponding kernel is
  \begin{multline*}
    \kkk((\mu,x),(\mu',x'))
    :=
    \left\langle
      \Phi(\mu,x),
      \Phi(\mu',x')
    \right\rangle_{\HHH}\\
    =
    \sum_{j=0}^{1}
    a_j
    \left\langle
      \Phi_j(\mu,x),
      \Phi_j(\mu',x')
    \right\rangle_{\HHH_j}
    =
    \sum_{j=0}^{1}
    a_j
    \kkk_j((\mu,x),(\mu',x')),
  \end{multline*}
  where $\kkk_0$ and $\kkk_1$ are the kernels corresponding to $\Phi_0$ and $\Phi_1$,
  respectively.
  It is clear that this kernel is forecast-continuous.
  Applying the ALN to it and using~(\ref{eq:ALN}),
  we obtain
  \begin{multline*}
    a_j
    \left\|
      \sum_{n=1}^N
      (y_n-\mu_n)
      \Phi_j(\mu_n,x_n)
    \right\|^2_{\HHH_j}\\
    \le
    \left\|
      \left(
        \sum_{n=1}^N
        (y_n-\mu_n)
        \Phi_0(\mu_n,x_n),
        \sum_{n=1}^N
        (y_n-\mu_n)
        \Phi_1(\mu_n,x_n)
      \right)
    \right\|^2_{\HHH}\\
    =
    \left\|
      \sum_{n=1}^N
      (y_n-\mu_n)
      \Phi(\mu_n,x_n)
    \right\|^2_{\HHH}
    \le
    Y^2
    \sum_{n=1}^N
    \left\|
      \Phi(\mu_n,x_n)
    \right\|^2_{\HHH}\\
    =
    Y^2
    \sum_{n=1}^N
    \sum_{j=0}^{1}
    a_j
    \left\|
      \Phi_j(\mu_n,x_n)
    \right\|^2_{\HHH_j}.
    \qedmath
  \end{multline*}
\end{proof}

Merging $\Phi_0(\mu,x)=\mu$ and $\Phi_1(\mu,x)=\kkk_x$ by Corollary \ref{cor:mixture1},
we obtain
\begin{multline}\label{eq:calibration1}
  \left|
    \sum_{n=1}^N
    (y_n-\mu_n)
    \mu_n
  \right|
  =
  \left\|
    \sum_{n=1}^N
    (y_n-\mu_n)
    \Phi_0(\mu_n,x_n)
  \right\|_{\bbbr}\\
  \le
  \frac{Y}{\sqrt{a_0}}
  \sqrt
  {
    \sum_{n=1}^N
    \bigl(
      a_0 \mu_n^2
      +
      a_1 \kkk(x_n,x_n)
    \bigr)
  }
\end{multline}
and,
using (\ref{eq:simple-resolution}),
\begin{multline}\label{eq:resolution1}
  \left|
    \sum_{n=1}^N
    (y_n-\mu_n)
    D(x_n)
  \right|
  \le
  \left\|
    \sum_{n=1}^N
    (y_n-\mu_n)
    \kkk_{x_n}
  \right\|_{\FFF}
  \left\|
    D
  \right\|_{\FFF}\\
  =
  \left\|
    \sum_{n=1}^N
    (y_n-\mu_n)
    \Phi_1(\mu_n,x_n)
  \right\|_{\FFF}
  \left\|
    D
  \right\|_{\FFF}\\
  \le
  \frac{Y}{\sqrt{a_1}}
  \left\|
    D
  \right\|_{\FFF}
  \sqrt
  {
    \sum_{n=1}^N
    \bigl(
      a_0 \mu_n^2
      +
      a_1 \kkk(x_n,x_n)
    \bigr)
  },
\end{multline}
for each function $D\in\FFF$.

\subsection*{Proof proper}

The proof is based on the elementary inequality
\begin{align}
  &\sum_{n=1}^N
  (y_n-\mu_n)^2\notag\\
  &=
  \sum_{n=1}^N
  (y_n-D(x_n))^2
  +
  2
  \sum_{n=1}^N
  (D(x_n)-\mu_n)
  (y_n-\mu_n)
  -
  \sum_{n=1}^N
  (D(x_n)-\mu_n)^2\notag\\
  &\le
  \sum_{n=1}^N
  (y_n-D(x_n))^2
  +
  2
  \sum_{n=1}^N
  (D(x_n)-\mu_n)
  (y_n-\mu_n)\label{eq:fundamental}
\end{align}
(the intermediate equality follows from $a^2=(a-b)^2+2ab-b^2$).
Using this inequality and (\ref{eq:calibration1})--(\ref{eq:resolution1}),
we obtain for the $\mu_n\in[-Y,Y]$
output by the ALN with the merged kernel as parameter:
\begin{align*}
  &\sum_{n=1}^N
  (y_n-\mu_n)^2\\
  &\le
  \sum_{n=1}^N
  (y_n-D(x_n))^2
  +
  2
  \left|
    \sum_{n=1}^N
    \mu_n
    (y_n-\mu_n)
  \right|
  +
  2
  \left|
    \sum_{n=1}^N
    D(x_n)
    (y_n-\mu_n)
  \right|\\
  &\le
  \sum_{n=1}^N
  (y_n-D(x_n))^2
  +
  2 Y
  \left(
    \frac{1}{\sqrt{a_1}} \left\|D\right\|_{\FFF} + \frac{1}{\sqrt{a_0}}
  \right)
  \sqrt
  {
    \sum_{n=1}^N
    \bigl(
      a_0 \mu_n^2 + a_1 \kkk(x_n,x_n)
    \bigr)
  }\\
  &\le
  \sum_{n=1}^N
  (y_n-D(x_n))^2
  +
  2 Y
  \left(
    \frac{1}{\sqrt{a_1}} \left\|D\right\|_{\FFF} + \frac{1}{\sqrt{a_0}}
  \right)
  \sqrt{a_0 Y^2 + a_1 \ccc_{\FFF}^2}
  \sqrt{N}.
\end{align*}
It remains to set $a_1:=1$ and $a_0:=1/Y^2$.

\section{Proof of Theorem~\ref{thm:main2}}
\label{sec:proof2}

In this section we will modify
(essentially, further simplify)
the proof of Theorem \ref{thm:main1} given in the previous section
to obtain the proof of Theorem \ref{thm:main2}.

\subsection*{K29 algorithm}

A kernel $\kkk$ on $[-Y,Y]\times\mathbf{X}$ is \emph{K29-admissible}
if the function $\kkk((\mu,x),(\mu',x'))$ is continuous in $\mu\in[-Y,Y]$
for all fixed $\mu'\in[-Y,Y]$, $x\in\mathbf{X}$, and $x'\in\mathbf{X}$.
For such a kernel the function
\begin{equation}\label{eq:function2}
  S_n(\mu)
  :=
  \sum_{i=1}^{n-1}
  \kkk
  \bigl(
    (\mu,x_n),(\mu_i,x_i)
  \bigr)
  (y_i-\mu_i)
\end{equation}
is continuous in $\mu\in[-Y,Y]$.

\bigskip

\noindent
\textsc{The K29 algorithm}

\noindent
\textbf{Parameter:} K29-admissible kernel $\kkk$ on $[-Y,Y]\times\mathbf{X}$

\parshape=7
\IndentI   \WidthI
\IndentII  \WidthII
\IndentII  \WidthII
\IndentII  \WidthII
\IndentIII \WidthIII
\IndentII  \WidthII
\IndentI   \WidthI
\noindent
FOR $n=1,2,\dots$:\\
  Read $x_n\in\mathbf{X}$.\\
  Define $S_n:[-Y,Y]\to\bbbr$ by (\ref{eq:function2}).\\
  Output any root $\mu\in[-Y,Y]$ of $S_n(\mu)=0$ as $\mu_n$;\\
    if there are no roots, set $\mu_n:=Y\sign S_n$.\\
  Read $y_n\in[-Y,Y]$.\\
END FOR.

\bigskip

\noindent
Let us say that a feature mapping $\Phi(\mu,x)$ is \emph{K29-admissible}
if the kernel $\kkk$ defined by~(\ref{eq:K}) is K29-admissible.
\begin{theorem}\label{thm:K29}
  Let $\kkk$ be the kernel
  defined by~(\ref{eq:K})
  for a K29-admissible feature mapping $\Phi:[-Y,Y]\times\mathbf{X}\to\HHH$.
  The K29 algorithm with parameter $\kkk$ outputs $\mu_n\in[-Y,Y]$
  such that
  \begin{equation}\label{eq:K29}
    \left\|
      \sum_{n=1}^N
      (y_n-\mu_n)
      \Phi(\mu_n,x_n)
    \right\|_{\HHH}^2
    \le
    \sum_{n=1}^N
    (y_n-\mu_n)^2
    \left\|
      \Phi(\mu_n,x_n)
    \right\|_{\HHH}^2
  \end{equation}
  always holds for all $N=1,2,\dots$.
\end{theorem}
\begin{proof}
  Following the K29 algorithm
  Predictor ensures that Skeptic will never increase his capital
  with the strategy
  \begin{equation*}
    s_n
    :=
    \sum_{i=1}^{n-1}
    \kkk
    \bigl(
      (\mu_n,x_n),(\mu_i,x_i)
    \bigr)
    (y_i-\mu_i),
  \end{equation*}
  which implies
  \begin{align*}
    0
    &\ge
    \K_N-\K_0
    =
    \sum_{n=1}^N
    s_n(y_n-\mu_n)\\
    &=
    \sum_{n=1}^N
    \sum_{i=1}^{n-1}
    \kkk
    \bigl(
      (\mu_n,x_n),(\mu_i,x_i)
    \bigr)
    (y_n-\mu_n)
    (y_i-\mu_i)\\
    &=
    \frac12
    \sum_{n=1}^N
    \sum_{i=1}^N
    \kkk
    \bigl(
      (\mu_n,x_n),(\mu_i,x_i)
    \bigr)
    (y_n-\mu_n)
    (y_i-\mu_i)\\
    &\quad{}-
    \frac12
    \sum_{n=1}^N
    \kkk
    \bigl(
      (\mu_n,x_n),(\mu_n,x_n)
    \bigr)
    (y_n-\mu_n)^2\\
    &=
    \frac12
    \left\|
      \sum_{n=1}^N
      (y_n-\mu_n)
      \Phi(\mu_n,x_n)
    \right\|_{\HHH}^2
    -
    \frac12
    \sum_{n=1}^N
    (y_n-\mu_n)^2
    \left\|
      \Phi(\mu_n,x_n)
    \right\|_{\HHH}^2,
  \end{align*}
  which in turn implies~(\ref{eq:K29}).
  \qedtext
\end{proof}

\subsection*{Mixing feature mappings}

Now we have the following corollary
of Theorem \ref{thm:K29}.
\begin{corollary}\label{cor:mixture2}
  Let $\Phi_j:[-Y,Y]\times\mathbf{X}\to\HHH_j$, $j=0,1$,
  be forecast-continuous mappings from $[-Y,Y]\times\mathbf{X}$
  to Hilbert spaces $\HHH_j$,
  and let $a_j$, $j=0,1$, be positive constants.
  The forecasts $\mu_n\in[-Y,Y]$ output by the K29 algorithm
  with a suitable kernel parameter always satisfy
  \begin{multline*}
    \left\|
      \sum_{n=1}^N
      (y_n-\mu_n)
      \Phi_j(\mu_n,x_n)
    \right\|^2_{\HHH_j}\\
    \le
    \frac{1}{a_j}
    \sum_{n=1}^N
    (y_n-\mu_n)^2
    \left(
      a_0
      \left\|
        \Phi_0(\mu_n,x_n)
      \right\|^2_{\HHH_0}
      +
      a_1
      \left\|
        \Phi_1(\mu_n,x_n)
      \right\|^2_{\HHH_1}
    \right)
  \end{multline*}
  for all $N$ and for both $j=0$ and $j=1$.
\end{corollary}
\begin{proof}
  Being forecast-continuous,
  the kernel $\kkk$ defined in the proof of Corollary \ref{cor:mixture1}
  is \emph{a fortiori} K29-admissible.
  Applying the K29 algorithm to it and using~(\ref{eq:K29}),
  we obtain
  \begin{multline*}
    a_j
    \left\|
      \sum_{n=1}^N
      (y_n-\mu_n)
      \Phi_j(\mu_n,x_n)
    \right\|^2_{\HHH_j}\\
    \le
    \left\|
      \sum_{n=1}^N
      (y_n-\mu_n)
      \Phi(\mu_n,x_n)
    \right\|^2_{\HHH}
    \le
    \sum_{n=1}^N
    (y_n-\mu_n)^2
    \left\|
      \Phi(\mu_n,x_n)
    \right\|^2_{\HHH}\\
    =
    \sum_{n=1}^N
    (y_n-\mu_n)^2
    \sum_{j=0}^{1}
    a_j
    \left\|
      \Phi_j(\mu_n,x_n)
    \right\|^2_{\HHH_j}.
    \qedmath
  \end{multline*}
\end{proof}

Merging $\Phi_0(\mu,x)=\mu$ and $\Phi_1(\mu,x)=\kkk_x$ by Corollary \ref{cor:mixture2},
we obtain
\begin{multline}\label{eq:calibration2}
  \left|
    \sum_{n=1}^N
    (y_n-\mu_n)
    \mu_n
  \right|
  =
  \left\|
    \sum_{n=1}^N
    (y_n-\mu_n)
    \Phi_0(\mu_n,x_n)
  \right\|_{\bbbr}\\
  \le
  \sqrt
  {
    \frac{1}{a_0}
    \sum_{n=1}^N
    (y_n-\mu_n)^2
    \bigl(
      a_0 \mu_n^2
      +
      a_1 \kkk(x_n,x_n)
    \bigr)
  }\\
  \le
  \frac{1}{\sqrt{a_0}}
  \sqrt
  {
    a_1 \ccc_{\FFF}^2
    +
    a_0 Y^2
  }
  \sqrt
  {
    \sum_{n=1}^N
    (y_n-\mu_n)^2
  }
\end{multline}
and,
using (\ref{eq:simple-resolution}),
\begin{multline}\label{eq:resolution2}
  \left|
    \sum_{n=1}^N
    (y_n-\mu_n)
    D(x_n)
  \right|
  \le
  \left\|
    \sum_{n=1}^N
    (y_n-\mu_n)
    \kkk_{x_n}
  \right\|_{\FFF}
  \left\|
    D
  \right\|_{\FFF}\\
  =
  \left\|
    \sum_{n=1}^N
    (y_n-\mu_n)
    \Phi_1(\mu_n,x_n)
  \right\|_{\FFF}
  \left\|
    D
  \right\|_{\FFF}\\
  \le
  \left\|
    D
  \right\|_{\FFF}
  \sqrt
  {
    \frac{1}{a_1}
    \sum_{n=1}^N
    (y_n-\mu_n)^2
    \bigl(
      a_0 \mu_n^2
      +
      a_1 \kkk(x_n,x_n)
    \bigr)
  }\\
  \le
  \frac{1}{\sqrt{a_1}}
  \left\|
    D
  \right\|_{\FFF}
  \sqrt
  {
    a_1 \ccc_{\FFF}^2
    +
    a_0 Y^2
  }
  \sqrt
  {
    \sum_{n=1}^N
    (y_n-\mu_n)^2
  },
\end{multline}
for each function $D\in\FFF$.

\subsection*{Proof proper}

Using (\ref{eq:fundamental}) and (\ref{eq:calibration2})--(\ref{eq:resolution2})
with $a_0:=a$ and $a_1:=1$,
we obtain for the $\mu_n$ output by the K29 algorithm
with the merged kernel as parameter:
\begin{align*}
  &\sum_{n=1}^N
  (y_n-\mu_n)^2\\
  &\le
  \sum_{n=1}^N
  (y_n-D(x_n))^2
  +
  2
  \left|
    \sum_{n=1}^N
    \mu_n
    (y_n-\mu_n)
  \right|
  +
  2
  \left|
    \sum_{n=1}^N
    D(x_n)
    (y_n-\mu_n)
  \right|\\
  &\le
  \sum_{n=1}^N
  (y_n-D(x_n)^2
  +
  2
  \sqrt
  {
    \ccc_{\FFF}^2 + a Y^2
  }
  \left(
    \left\|D\right\|_{\FFF} + \frac{1}{\sqrt{a}}
  \right)
  \sqrt
  {
    \sum_{n=1}^N
    (y_n-\mu_n)^2
  }.
\end{align*}
The inequality between the extreme terms of this chain
is quadratic in
\begin{equation*}
  \sqrt
  {
    \sum_{n=1}^N
    (y_n-\mu_n)^2
  };
\end{equation*}
solving it,
we obtain
\begin{multline*}
  \sqrt
  {
    \sum_{n=1}^N
    (y_n-\mu_n)^2
  }
  \le
  \sqrt
  {
    \sum_{n=1}^N
    (y_n - D(x_n))^2
    +
    \left(
      \ccc_{\FFF}^2 + a Y^2
    \right)
    \left(
      \left\|D\right\|_{\FFF} + \frac{1}{\sqrt{a}}
    \right)^2
  }\\
  +
  \sqrt
  {
    \ccc_{\FFF}^2 + a Y^2
  }
  \left(
    \left\|D\right\|_{\FFF} + 1/\sqrt{a}
  \right),
\end{multline*}
which is equivalent to (\ref{eq:goal2}) when $a=1/Y^2$.

\section{Bayes-type competitive on-line regression
and proof of Theorem~\ref{thm:main3}}
\label{sec:proof3}

The first result in the Bayes-style competitive on-line regression
appears to be the following:
if the benchmark class $\FFF$ consists of the linear functions
$D(x)=\left\langle\theta,x\right\rangle$ on $\mathbf{X}=\bbbr^m$
whose ``complexity'' is measured by the $L^2$ norm
$\left\|\theta\right\|_2:=\sqrt{\sum_{i=1}^m\theta_i^2}$
of $\theta$'s components $\theta_i$
and if $a$ is a positive constant,
some on-line prediction algorithm (namely, the Aggregating Algorithm) ensures
\begin{align}
  &\sum_{n=1}^N
  \left(
    y_n - \mu_n
  \right)^2
  \notag\\
  &\le
  \sum_{n=1}^N
  \left(
    y_n - \left\langle\theta,x_n\right\rangle
  \right)^2
  +
  a
  \left\|
    \theta
  \right\|_2^2
  +
  Y^2
  \ln\det
  \left(
    I
    +
    \frac{1}{a}
    \sum_{n=1}^N
    x_n x'_n
  \right)
  \label{eq:AAR}\\
  &\le
  \sum_{n=1}^N
  \left(
    y_n - \left\langle\theta,x_n\right\rangle
  \right)^2
  +
  a
  \left\|
    \theta
  \right\|_2^2
  +
  Y^2
  \sum_{i=1}^m
  \ln
  \left(
    1
    +
    \frac{1}{a}
    \sum_{n=1}^N
    x_{n,i}^2
  \right),
  \notag
\end{align}
for all $N$ and all $\theta\in\bbbr^m$
(\cite{vovk:2001competitive}, Theorem 1;
different proofs are given in \cite{azoury/warmuth:2001}, Theorem 4.6,
and \cite{forster:1999}).
In particular, if $\left\|\theta\right\|_2$ and all components $x_{n,i}$ of all $x_n$
are bounded by a constant,
\begin{equation*}
  \sum_{n=1}^N
  \left(
    y_n - \mu_n
  \right)^2
  \le
  \sum_{n=1}^N
  \left(
    y_n - \left\langle\theta,x_n\right\rangle
  \right)^2
  +
  O
  \left(
    \ln N
  \right);
\end{equation*}
it is interesting that the regret term is now $O(\ln N)$,
rather than $O(\surd N)$ as in (\ref{eq:goal1}).

\ifFULL\bluebegin
  The claim of \cite{vovk:2001competitive} about the RR algorithm might appear wrong
  (as noticed by Manfred Warmuth, who discovered an actual error
  in an early version of that paper,
  and Tong Zhang)
  but is in fact right:
  notice that the prediction $\mathrm{RR}^Y$ is \emph{not} the truncated RR prediction:
  it is the experts' predictions that get truncated.

  In the finite-dimensional case the typical regret is of the order of magnitude $\ln N$;
  this agrees with the statistical-learning result by \cite{lee/etal:1998}
  that $\epsilon\propto 1/N$ when the benchmark class is convex
  and has finite pseudo-dimension
  (indeed, $\sum_{n=1}^N 1/n \approx \ln N$).
\blueend\fi

We are, however,
interested in the infinite-dimensional benchmark classes.
The result (\ref{eq:AAR}) was carried over to separable RKHS
in \cite{gammerman/etal:2004UAI}:
there is an on-line prediction algorithm that ensures
\begin{equation}\label{eq:Yura}
  \sum_{n=1}^N
  \left(
    y_n - \mu_n
  \right)^2
  \le
  \sum_{n=1}^N
  \left(
    y_n - D(x_n)
  \right)^2
  +
  a
  \left\|
    D
  \right\|_{\FFF}^2
  +
  Y^2
  \ln\det
  \left(
    I
    +
    \frac{1}{a}
    K
  \right)
\end{equation}
for all $N$ and all prediction rules $D$ in a separable RKHS $\FFF$ on $\mathbf{X}$,
where $K$ is the $N\times N$ \emph{Gram matrix}
with the elements $K_{i,j}:=\kkk(x_i,x_j)$, $i,j=1,\ldots,N$,
and $\kkk$ is $\FFF$'s \ifCONF``\fi{}reproducing kernel\ifCONF''\fi.
(Actually this result is stated in \cite{gammerman/etal:2004UAI}
only for prediction rules $D$ of the form
$\sum_{i=1}^k c_i\kkk_{z_i}$,
where $k\in\{1,2,\ldots\}$,
$c_1,\ldots,c_k\in\bbbr$, and $z_1,\ldots,z_k\in\mathbf{X}$;
but the result is true in general since such prediction rules
are dense in $\FFF$:
see \cite{aronszajn:1950}, \S I.2, (4).
Alternatively, the general result follows by the representer theorem,
stated in, e.g., \cite{kimeldorf/wahba:1971} and \cite{scholkopf/smola:2002},
Theorem 4.2 on p.~90.)

A disadvantage of the bound (\ref{eq:Yura}) is that,
for a fixed $a$,
the term
\begin{equation*}
  \ln\det
  \left(
    I
    +
    \frac{1}{a}
    K
  \right)
\end{equation*}
(which also occurs in \cite{kakade/etal:2005}, Theorems 3.1 and 3.2,
and \cite{cesabianchi/etal:2002})
can have order of magnitude $N$:
indeed, if
\begin{equation*}
  \kkk(x_i,x_j)
  =
  \begin{cases}
    1 & \text{if $i=j$}\\
    0 & \text{otherwise},
  \end{cases}
\end{equation*}
this term becomes
\begin{equation*}
  \ln
  \prod_{n=1}^N
  \left(
    1
    +
    \frac{1}{a}
  \right)
  =
  N
  \ln(1+1/a).
\end{equation*}
More generally,
Minkowski's result from \cite{beckenbach/bellman:1965}, Chapter 2, Theorem 15,
shows that
\begin{equation*}
  \ln\det
  \left(
    I
    +
    \frac{1}{a}
    K
  \right)
  \ge
  N
  \ln
  \left(
    1
    +
    \det{}^{1/N}
    \left(
      \frac{1}{a}
      K
    \right)
  \right),
\end{equation*}
and so this term will not be small as compared to $N$
unless $\det K \le (a\epsilon)^N$
for a small $\epsilon>0$.

Our argument in the previous paragraph assumed that $a$ was fixed.
Let us now see what (\ref{eq:Yura}) leads to when $N$
and an upper bound $d$ on $\left\|D\right\|_{\FFF}$
are given in advance,
which gives some scope for optimizing $a$.
In conjunction with the fact that the determinant of a positive definite matrix
does not exceed the product of its diagonal elements
(\cite{beckenbach/bellman:1965}, Chapter 2, Theorem 7),
(\ref{eq:Yura}) implies
\begin{multline}\label{eq:YuraSimple}
  \sum_{n=1}^N
  \left(
    y_n - \mu_n
  \right)^2
  \le
  \sum_{n=1}^N
  \left(
    y_n - D(x_n)
  \right)^2
  +
  a
  \left\|
    D
  \right\|_{\FFF}^2
  +
  Y^2
  N
  \ln
  \left(
    1
    +
    \frac{\ccc_{\FFF}^2}{a}
  \right)\\
  \le
  \sum_{n=1}^N
  \left(
    y_n - D(x_n)
  \right)^2
  +
  a d^2
  +
  \frac{Y^2\ccc_{\FFF}^2N}{a}.
\end{multline}
The minimum of $ad^2+Y^2\ccc_{\FFF}^2N/a$ is achieved at $a=(Y\ccc_{\FFF}/d)\surd N$,
and for this value of $a$ (\ref{eq:YuraSimple}) becomes
\begin{equation*}
  \sum_{n=1}^N
  \left(
    y_n - \mu_n
  \right)^2
  \le
  \sum_{n=1}^N
  \left(
    y_n - D(x_n)
  \right)^2
  +
  2Y\ccc_{\FFF}d\sqrt{N}.
\end{equation*}
We can see an analogue of the familiar term
$2Y\ccc_{\FFF}\left\|D\right\|_{\FFF}\surd N$.
The expression
\begin{equation*}
  \frac{3}{2}
  Y^2
  \ln N
  +
  \frac{\ccc_{\FFF}^2\left\|D\right\|_{\FFF}^2}{4}
  +
  O(Y^2)
\end{equation*}
in (\ref{eq:goal3})
can be interpreted as the price that we pay for not knowing
$\left\|D\right\|_{\FFF}$ and $N$ in advance.

\subsection*{Kummer's $U$ function}

In the proof of Theorem \ref{thm:main3}
we will need an approximation to Kummer's $U$ function
\begin{equation}\label{eq:U}
  U(a,b,z)
  :=
  \frac{1}{\Gamma(a)}
  \int_0^{\infty}
  e^{-zt}
  t^{a-1}
  (1+t)^{b-a-1}
  dt
\end{equation}
from \cite{taylor:1939}, p.~44, (26a).
A concise statement of this result is given
in \cite{erdelyi:1953}, p.~281, \S6.13.3, (22)
(in fact, Taylor states his result
in terms of the closely related Whittaker function;
\cite{erdelyi:1953} states it in terms of Kummer's $U$ function,
which is, however, denoted $\Psi$:
cf.\ (2) on p.~255;
in using the notation $U$ we are following
\cite{abramowitz/stegun:1964}, Chapter 13:
cf.\ 13.2.5 on p.~505).
The approximation is given by the formula
\begin{multline}\label{eq:approximation}
  \kappa^{-\kappa}
  z^{b/2-1/4}
  (z-4\kappa)^{1/4}
  e^{\kappa-z/2}
  U(a,b,z)\\
  =
  e^{i\xi}
  \left(
    1
    +
    O
    \left(
      \left|
        \kappa
      \right|^{-r}
    \right)
    +
    O
    \left(
      \left|
        \xi
      \right|^{-1}
    \right)
  \right),
\end{multline}
where $r\in(0,1]$,
\begin{align*}
  \kappa
  &:=
  b/2-a,\\
  i\xi
  &:=
  \kappa
  \ln
  \frac
  {
    \left(
      z^{1/2}
      +
      (z-4\kappa)^{1/2}
    \right)^2
  }
  {4\kappa}
  -
  \frac12
  z^{1/2}
  (z-4\kappa)^{1/2},
\end{align*}
and it is assumed that $\xi\to\infty$
and
\begin{equation}\label{eq:r}
  \lvert z\rvert
  >
  \delta\lvert\kappa\rvert^{-1+2r}
\end{equation}
for some constant $\delta>0$.
We are only interested in the case
$z>0$, $a\ge1$, and $b\in[0,1]$,
which also implies $\kappa<0$.
Since $\ln(-1)=\pm i\pi$,
the expression for $i\xi$ can be rewritten as
\begin{equation}\label{eq:ixi}
  i\xi
  =
  \kappa
  \ln
  \frac
  {
    \left(
      z^{1/2}
      +
      (z+4\lvert\kappa\rvert)^{1/2}
    \right)^2
  }
  {4\lvert\kappa\rvert}
  -
  \frac12
  z^{1/2}
  (z+4\lvert\kappa\rvert)^{1/2}
  \pm
  \kappa i \pi,
\end{equation}
and we can see that $\Re(i\xi)<0$ and, therefore, $\arg\xi\in(0,\pi)$;
we have already used this fact in choosing the expression (\ref{eq:approximation})
among the expressions given in \cite{erdelyi:1953}, p.~281, \S6.13.3
((21)--(24)).
Using (\ref{eq:ixi}) and the fact that $\lvert\xi\rvert\ge\lvert\kappa\rvert\pi$,
we deduce from (\ref{eq:approximation}):
\begin{multline*}
  \ln U(a,b,z)
  =
  \kappa \ln\kappa
  +
  \left(
    \frac14 - \frac{b}{2}
  \right)
  \ln z
  -
  \frac14
  \ln(z-4\kappa)
  -
  \kappa
  +
  \frac{z}{2}\\
  +
  i\xi
  +
  O
  \left(
    \left|
      \kappa
    \right|^{-r}
  \right)\\
  =
  \kappa \ln\lvert\kappa\rvert
  +
  \left(
    \frac14 - \frac{b}{2}
  \right)
  \ln z
  -
  \frac14
  \ln(z-4\kappa)
  -
  \kappa
  +
  \frac{z}{2}\\
  -
  \frac12
  z^{1/2}
  (z-4\kappa)^{1/2}
  +
  \kappa
  \ln
  \frac
  {
    \left(
      z^{1/2}
      +
      (z+4\lvert\kappa\rvert)^{1/2}
    \right)^2
  }
  {4\lvert\kappa\rvert}
  +
  O
  \left(
    \left|
      \kappa
    \right|^{-r}
  \right)\\
  =
  \left(
    \frac14 - \frac{b}{2}
  \right)
  \ln z
  -
  \frac14
  \ln
  \left(
    \frac{z}{4} - \kappa
  \right)
  -
  \frac12
  \ln 2
  -
  \kappa
  +
  \frac{z}{2}\\
  -
  z^{1/2}
  \left(
    \frac{z}{4} - \kappa
  \right)^{1/2}
  +
  2
  \kappa
  \ln
  \left(
    \left(
      \frac{z}{4}
    \right)^{1/2}
    +
    \left(
      \frac{z}{4} - \kappa
    \right)^{1/2}
  \right)
  +
  O
  \left(
    \left|
      \kappa
    \right|^{-r}
  \right)\\
  =
  \left(
    \frac14 - \frac{b}{2}
  \right)
  \ln z
  -
  \frac14
  \ln
  \left(
    a - \frac{b}{2} + \frac{z}{4}
  \right)
  -
  \frac12
  \ln 2
  +
  a
  -
  \frac{b}{2}
  +
  \frac{z}{2}\\
  -
  z^{1/2}
  \left(
    a - \frac{b}{2} + \frac{z}{4}
  \right)^{1/2}
  +
  2
  \left(
    \frac{b}{2} - a
  \right)
  \ln
  \left(
    \left(
      \frac{z}{4}
    \right)^{1/2}
    +
    \left(
      a - \frac{b}{2} + \frac{z}{4}
    \right)^{1/2}
  \right)\\
  +
  O
  \left(
    \left|
      \kappa
    \right|^{-r}
  \right).
\end{multline*}
By Stirling's formula
(\cite{abramowitz/stegun:1964}, p.~257, 6.1.41),
\begin{equation*}
  \ln\Gamma(a)
  =
  -a
  +
  \left(
    a-\frac12
  \right)
  \ln a
  +
  \frac12\ln(2\pi)
  +
  O
  \left(
    a^{-1}
  \right),
\end{equation*}
which for $b\in[0,1]$ gives
\begin{multline}\label{eq:ln}
  -\ln
  \left(
    \Gamma(a)
    U(a,b,z)
  \right)
  =
  \left(
    \frac{b}{2} - \frac14
  \right)
  \ln z
  +
  \frac14
  \ln
  \left(
    a - \frac{b}{2} + \frac{z}{4}
  \right)
  +
  \frac{b}{2}
  -
  \frac{z}{2}\\
  +
  z^{1/2}
  \left(
    a - \frac{b}{2} + \frac{z}{4}
  \right)^{1/2}
  +
  2
  \left(
    a - \frac{b}{2}
  \right)
  \ln
  \left(
    \left(
      a - \frac{b}{2} + \frac{z}{4}
    \right)^{1/2}
    +
    \left(
      \frac{z}{4}
    \right)^{1/2}
  \right)\\
  -
  \left(
    a - \frac12
  \right)
  \ln a
  -
  \frac12
  \ln\pi
  +
  O
  \left(
    a^{-r}
  \right)\\
  =
  \left(
    \frac{b}{2} - \frac{1}{4}
  \right)
  \ln z
  +
  \frac14
  \ln
  \left(
    a - \frac{b}{2} + \frac{z}{4}
  \right)
  +
  \frac{b}{2}
  -
  \frac{z}{2}\\
  +
  z^{1/2}
  \left(
    a - \frac{b}{2} + \frac{z}{4}
  \right)^{1/2}
  +
  2
  \left(
    a - \frac{b}{2}
  \right)
  \ln
  \left(
    \left(
      1-\frac{b}{2a}+\frac{z}{4a}
    \right)^{1/2}
    +
    \left(
      \frac{z}{4a}
    \right)^{1/2}
  \right)\\
  +
  \left(
    \frac12 - \frac{b}{2}
  \right)
  \ln a
  -
  \frac12
  \ln\pi
  +
  O
  \left(
    a^{-r}
  \right).
\end{multline}

\subsection*{Proof of Theorem~\ref{thm:main3}}

To get rid of the parameter $a$ in the first inequality of (\ref{eq:YuraSimple}),
we will merge the AA predictions
(truncated to $[-Y,Y]$ if necessary\ifFULL\bluebegin\ [perhaps this is never needed:
see \cite{vovk:2001competitive}\blueend\fi)
corresponding to all possible $a\in(0,\infty)$
w.r.\ to the probability measure
\begin{equation*}
  Q(da)
  :=
  \frac{\epsilon c^{2\epsilon}}{(a+c^2)^{1+\epsilon}}
  da
\end{equation*}
on $(0,\infty)$;
here and in what follows we let $c$ stand for $\ccc_{\FFF}$
and $\epsilon$ for a constant in $(0,1]$ to be chosen later.
Taking $\eta:=1/(2Y^2)$
(see \cite{vovk:2001competitive}, towards the end of \S2.4)
and $\beta:=e^{-\eta}$,
making use of Lemmas 1 and 2 of \cite{vovk:2001competitive},
and letting $d$ stand for $\left\|D\right\|_{\FFF}$,
we obtain the following bound
for the excess loss of the merged predictions over $D$'s predictions
over the first $N$ rounds:
\begin{multline*}
  \log_{\beta}
  \int
    \beta^{ad^2+Y^2N\ln(1+c^2/a)}
  Q(da)
  =
  -\frac{1}{\eta}
  \ln
  \int
  e^{-\eta a d^2}
  \left(
    1+\frac{c^2}{a}
  \right)^{-\eta Y^2 N}
  Q(da)\\
  =
  -2Y^2
  \ln
  \int
  e^{-a d^2/(2Y^2)}
  \left(
    1+\frac{c^2}{a}
  \right)^{-N/2}
  Q(da)\\
  =
  -2Y^2
  \ln
  \left(
    \epsilon c^{2\epsilon}
    \int_0^{\infty}
    e^{-a d^2/(2Y^2)}
    \left(
      a+c^2
    \right)^{-N/2-1-\epsilon}
    a^{N/2}
    da
  \right).
\end{multline*}
Substituting $c^2t$ for $a$ transforms this to
\begin{equation*}
  -2Y^2
  \ln
  \left(
    \epsilon
    \int_0^{\infty}
    e^{-c^2d^2t/(2Y^2)}
    \left(
      1+t
    \right)^{-N/2-1-\epsilon}
    t^{N/2}
    dt
  \right),
\end{equation*}
which, by (\ref{eq:U}) and (\ref{eq:ln}), can be written as
\begin{multline}\label{eq:clumsy}
  -2Y^2
  \ln
  \left(
    \epsilon
    \Gamma
    \left(
      \frac{N}{2}+1
    \right)
    U
    \left(
      \frac{N}{2}+1,
      1-\epsilon,
      \frac{c^2d^2}{2Y^2}
    \right)
  \right)\\
  =
  -2Y^2\ln\epsilon
  +
  Y^2
  \left(
    \frac12 - \epsilon
  \right)
  \ln\frac{c^2d^2}{2Y^2}
  +
  \frac{Y^2}{2}
  \ln
  \left(
    \frac{N}{2} + \frac12 + \frac{\epsilon}{2} + \frac{c^2d^2}{8Y^2}
  \right)\\
  +
  Y^2
  (1-\epsilon)
  -
  \frac{c^2d^2}{2}
  +
  Ycd
  \left(
    N + 1 + \epsilon + \frac{c^2d^2}{4Y^2}
  \right)^{1/2}\\
  +
  2Y^2
  (N+1+\epsilon)
  \ln
  \left(
    \left(
      1
      -
      \frac{1-\epsilon}{N+2}
      +
      \frac{c^2d^2}{4Y^2(N+2)}
    \right)^{1/2}
    +
    \frac{cd}{2Y\sqrt{N+2}}
  \right)\\
  +
  \epsilon Y^2
  \ln
  \left(
    \frac{N}{2} + 1
  \right)
  -
  Y^2\ln\pi
  +
  Y^2
  O
  \left(
    N^{-r}
  \right).
\end{multline}

Remember that the validity of the approximation (\ref{eq:approximation})
requires the condition (\ref{eq:r}).
In the most interesting case $1\ll z\ll\kappa$
we can take $r:=1/2$ to get the best bound,
corresponding to the accuracy of $O(N^{-1/2})$;
unfortunately, such a bound would be rather clumsy,
and the following transformations
(which sometimes are inequalities rather than equalities)
are performed to a much worse accuracy, $O(Y^2)$\Extra{ (another reason being
Grothers's remarks in \cite{grothers:1972}, p.~1681)}.
To make sure that (\ref{eq:r}) holds it is now sufficient
to assume that
\begin{equation}\label{eq:cdY}
  \frac{c^2d^2}{Y^2}
  \ge
  \delta^2 N^{-1+2\delta}
\end{equation}
for some constant $\delta$;
we will first assume that this condition holds,
and at the end of the proof will get rid of it
(although not completely:
it survives in the presence of ``$\max$'' in (\ref{eq:goal3})).

The fourth addend from the end of (\ref{eq:clumsy})
can be bounded above as follows:
\begin{multline*}
  2Y^2
  (N+1+\epsilon)
  \ln
  \left(
    \left(
      1
      -
      \frac{1-\epsilon}{N+2}
      +
      \frac{c^2d^2}{4Y^2(N+2)}
    \right)^{1/2}
    +
    \frac{cd}{2Y\sqrt{N+2}}
  \right)\\
  \le
  2Y^2
  (N+1+\epsilon)
  \ln
  \left(
    1
    +
    \frac{c^2d^2}{8Y^2(N+2)}
    +
    \frac{cd}{2Y\sqrt{N+2}}
  \right)\\
  \le
  2Y^2
  (N+1+\epsilon)
  \left(
    \frac{c^2d^2}{8Y^2(N+2)}
    +
    \frac{cd}{2Y\sqrt{N+2}}
  \right)\\
  \le
  \frac{c^2d^2}{4}
  +
  Ycd
  \sqrt{N+1+\epsilon}.
\end{multline*}
This allows us to bound (\ref{eq:clumsy}) from above by
\begin{multline*}
  (1-2\epsilon)
  Y^2
  \ln\frac{cd}{Y}
  +
  \frac{Y^2}{2}
  \ln
  \left(
    N + \frac{c^2d^2}{Y^2}
  \right)
  -
  \frac{c^2d^2}{2}
  +
  Ycd
  \left(
    N + 1 + \epsilon + \frac{c^2d^2}{4Y^2}
  \right)^{1/2}\\
  +
  \frac{c^2d^2}{4}
  +
  Ycd
  \sqrt{N+1+\epsilon}
  +
  \epsilon Y^2
  \ln N
  +
  O
  \left(
    Y^2
  \right)\\
  \le
  (2-2\epsilon)
  Y^2
  \ln\frac{cd}{Y}
  +
  \left(
    \frac12 + \epsilon
  \right)
  Y^2
  \ln N
  +
  2Ycd
  \sqrt{N+1+\epsilon}
  +
  \frac{c^2d^2}{4}
  +
  O
  \left(
    Y^2
  \right).
\end{multline*}
If we take $\epsilon=1$,
this will give
\begin{equation}\label{eq:last-equation}
  \frac32
  Y^2
  \ln N
  +
  2Ycd
  \sqrt{N+2}
  +
  \frac{(cd)^2}{4}
  +
  O
  \left(
    Y^2
  \right),
\end{equation}
i.e., (\ref{eq:goal3}).
The choice of $\epsilon=1$ appears to lead to the simplest regret term,
but notice that by choosing $\epsilon$ close to $0$
we can improve the constant $\frac32Y^2$
in the second leading addend in the regret term in (\ref{eq:goal3})
making it close to $\frac12Y^2$.

Returning to the condition (\ref{eq:cdY}),
it is easy to check that the bound (\ref{eq:last-equation})
will remain valid without this condition
if $cd$ is replaced by
\begin{equation*}
  P
  :=
  \max
  \left(
    cd,
    Y\delta N^{-1/2+\delta}
  \right);
\end{equation*}
indeed, this immediately follows from the monotonicity of $\Gamma(a)U(a,b,z)$ in $z$.
It remains to notice that the difference
between the addend $(cd)^2/4$ in (\ref{eq:last-equation})
and $P^2/4$ can be accommodated in the $O(Y^2)$ term,
so this addend can be left as it is.

\ifFULL\bluebegin
According to Grothers \cite{grothers:1972} (p.~1681),
the results of Taylor \cite{taylor:1939} are suspicious,
and he claims, with a reference to the English translation of Buchholz \cite{buchholz:1953},
that Taylor's asymptotic forms are only an order of magnitude approximations.
\blueend\fi

\section{Proof of Theorem~\ref{thm:main4}}
\label{sec:proof4}

Let $\FFF$ be the RKHS corresponding to the kernel on $\mathbf{X}=\bbbr$
defined by
$\kkk(x,x'):=c^2h(x-x')$,
where $h:\bbbr\to\bbbr$ is the ``triangular'' function
$h(t):=\max(1-\lvert t\rvert,0)$;
the positive definiteness of $\kkk$ follows from Bochner's theorem
(see, e.g., \cite{feller:1971}, \S XIX.2)
and Polya's theorem
(\cite{feller:1971}, Example (b) in \S XV.3).
\ifFULL\bluebegin
  By \cite{steinwart:2001}, Corollary 11,
  $\FFF$ is a universal RKHS.
\blueend\fi
Representation (\ref{eq:equiv}) shows that $\ccc_{\FFF}=c$.

Reality's strategy is $x_n:=2n$ and $y_n:=\pm Y$,
with $\sign(y_n)$ opposite to $\sign(\mu_n)$
(when $\mu_n=0$, $\sign(y_n)$ is chosen arbitrarily).
This will make sure that the loss of the on-line prediction algorithm
over the first $N$ rounds is at least $Y^2 N$.

Let $f_n\in\FFF$ be the function defined by $f_n(x):=ch(x-2n)$, $n=1,2,\ldots$\,.
It is clear that the functions $f_n$, $n=1,2,\ldots$, are orthogonal
and $\left\|f_n\right\|_{\FFF}=1$.
Set $\alpha:=cd/(Y\surd N)$
and let the decision rule $D\in\FFF$ be defined by
\begin{equation*}
  D
  :=
  \alpha
  \sum_{n=1}^N
  \frac{y_n}{c}
  f_n;  
\end{equation*}
one of the conditions of the theorem ensures that
$\alpha\le1$ and, therefore, $D$ takes values in $[-Y,Y]$.
The loss of $D$ over the first $N$ rounds is
$(1-\alpha)^2Y^2N$
and the norm of $D$ is $\left\|D\right\|_{\FFF}=\alpha(Y/c)\surd N=d$.
We can see that the excess loss of the prediction algorithm as compared to $D$ is
\begin{equation*}
  Y^2N
  -
  (1-\alpha)^2Y^2N
  =
  (2\alpha-\alpha^2)Y^2N
  =
  (2-\alpha)Ycd\sqrt{N},
\end{equation*}
which completes the proof.

\section{The algorithms}
\label{sec:algorithms}

In this short section we extract the prediction strategies
achieving (\ref{eq:goal1}) and (\ref{eq:goal2})
from our proof of Theorems \ref{thm:main1} and \ref{thm:main2}.
Replacing in (\ref{eq:function1}) the kernel $\kkk((\mu,x),(\mu',x'))$
by the merged kernel
$\mu\mu'/Y^2+\left\langle\kkk_x,\kkk_{x'}\right\rangle_{\FFF}$,
we obtain
\begin{equation}\label{eq:S}
  S_n(\mu)
  =
  \sum_{i=1}^{n-1}
  \Bigl(
    \mu \mu_i / Y^2
    +
    \kkk(x_n,x_i)
  \Bigr)
  (y_i-\mu_i)
  -
  \Bigl(
    \mu^2 / Y^2
    +
    \kkk(x_n,x_n)
  \Bigr)
  \mu;
\end{equation}
this immediately leads to the following explicit description
for the on-line prediction algorithm
we used in the proof of Theorem \ref{thm:main1}.

\bigskip

\noindent
\textsc{An algorithm achieving (\ref{eq:goal1})}

\noindent
\textbf{Parameter:} the reproducing kernel $\kkk$ of $\FFF$

\parshape=7
\IndentI   \WidthI
\IndentII  \WidthII
\IndentII  \WidthII
\IndentII  \WidthII
\IndentIII \WidthIII
\IndentII  \WidthII
\IndentI   \WidthI
\noindent
FOR $n=1,2,\dots$:\\
  Read $x_n\in\mathbf{X}$.\\
  Define $S_n(\mu)$ by (\ref{eq:S}) for all $\mu\in[-Y,Y]$.\\
  Define $\mu_n$ as any root $\mu\in[-Y,Y]$ of $S_n(\mu)=0$;\\
    if there are no roots, set $\mu_n:=Y\sign S_n$.\\
  Read $y_n\in[-Y,Y]$.\\
END FOR.

\bigskip

To obtain an algorithm achieving (\ref{eq:goal2}),
it suffices to replace (\ref{eq:S}) by
\begin{equation*}
  S_n(\mu)
  =
  \sum_{i=1}^{n-1}
  \Bigl(
    \mu \mu_i / Y^2
    +
    \kkk(x_n,x_i)
  \Bigr)
  (y_i-\mu_i).
\end{equation*}
\fi

\ifFULL\bluebegin
\section{Further research}

Competitive on-line statistics so far has been ``pure decision making'';
an inferential element (such as calibration and resolution of \cite{\GTPXIV}
needs to be added.
The most promising directions appear to be:
\begin{itemize}
\item
  the case where no \emph{a priori} upper bound $Y$ on $\lvert y_n\rvert$ is known;
\item
  several-steps ahead forecasting
  (this might solve the ``puzzle of the iterated logarithm'');
\item
  find the minimal amount of randomization needed for non-convex loss functions.
\end{itemize}
\blueend\fi

\subsection*{Acknowledgments}

I am grateful to Nicol\`o Cesa-Bianchi, Alex Smola, and, especially, Olivier Bousquet
for useful comments.
This work was partially supported by MRC (grant S505/65) and the Royal Society.

\end{document}